\documentclass{article}

\usepackage[nonatbib,final]{neurips_2020}

\usepackage{microtype}
\usepackage{graphicx}
\usepackage{booktabs}
\usepackage{hyperref}

\usepackage[utf8]{inputenc}
\usepackage[T1]{fontenc}
\usepackage{adjustbox}
\usepackage[noend]{algorithmic}
\usepackage{algorithmic}
\usepackage{amsfonts}
\usepackage[font=scriptsize]{subfig}
\usepackage{amsmath}
\usepackage{amssymb}
\usepackage{balance}
\usepackage{bbm}
\usepackage{bm}
\usepackage{caption}
\usepackage{cite}
\usepackage{clipboard}
\usepackage{enumitem}
\usepackage{float}
\usepackage[bottom,flushmargin]{footmisc}
\usepackage{makecell}
\usepackage{mathbbol}
\usepackage{mathtools, nccmath}
\usepackage{mathrsfs}
\usepackage{microtype}
\usepackage{multirow}
\usepackage{nicefrac}
\usepackage{pifont}
\usepackage{relsize}
\usepackage{scalerel}
\usepackage{setspace}
\usepackage{stackengine}
\usepackage{tcolorbox}
\usepackage{thmtools}
\usepackage{thm-restate}
\usepackage{tikz}
\usepackage{url}
\usepackage{wrapfig}
\usepackage{xcolor}

\usepackage[normalem]{ulem}

\usepackage{contour}
\usepackage{ulem}

\contourlength{0.6pt}
\newcommand{\muline}[1]{
\uline{\phantom{#1}}
\llap{\contour{white}{#1}}
}

\usepackage{algorithm}

\DeclareSymbolFontAlphabet{\mathbb}{AMSb}
\DeclareSymbolFontAlphabet{\mathbbl}{bbold}

\definecolor{crimson}{rgb}{0.7,0.01,0.02}
\definecolor{fern}{rgb}{0.05,0.6,0.05}
\definecolor{prussian}{rgb}{0,0.08,0.45}
\definecolor{faintgray}{gray}{0.9}
\definecolor{faintblue}{rgb}{0,0.08,0.65}

\newcommand{\pix}{\kern 0.1em}
\newcommand{\pmm}{\kern 0.25em$\pm$\kern 0.15em}
\newcommand{\pms}{\kern 0.10em$\pm$\kern 0.05em}

\newcommand{\QED}{\hfill\raisebox{-0.5pt}{\scalebox{0.88}{$\square$}}}
\newcommand{\EOD}{\hfill\raisebox{-0.5pt}{\scalebox{0.88}{\rotatebox[origin=c]{45}{$\square$}}}}
\newcommand{\dayum}[1]{{#1\parfillskip=0pt\par}}

\newcommand{\spaceqq}{\kern 0.2em=}
\newcommand{\spaceapp}{\kern 0.2em\approx}
\newcommand{\spacegeqq}{\kern 0.2em\geq}
\newcommand{\spaceleqq}{\kern 0.2em\leq}

\newcommand{\Pie}{\Pi}
\newcommand{\Tau}{\mathcal{T}}
\newcommand{\para}{\scalerel*{\pix||}{\bigstar}}

\declaretheorem[name=Theorem]{retheorem}
\declaretheorem[name=Proposition,numberlike=retheorem]{reproposition}

\declaretheorem[name=Definition]{redefinition}

\let\OLDthebibliography\thebibliography

\renewcommand\thebibliography[1]{
  \OLDthebibliography{#1}
  \setlength{\itemsep}{3pt plus 1pt minus 1pt}
}

\usepackage{letltxmacro}

\newlength{\commentindent}
\makeatletter
\renewcommand{\algorithmiccomment}[1]{\unskip\hfill\makebox[\commentindent][l]{$\triangleright$~#1}\par}
\LetLtxMacro{\oldalgorithmic}{\algorithmic}
\renewcommand{\algorithmic}[1][0]{%
  \oldalgorithmic[#1]%
  \renewcommand{\ALC@com}[1]{%
    \ifnum\pdfstrcmp{##1}{default}=0\else\algorithmiccomment{##1}\fi}%
}
\makeatother

\newif\ifdropfigures
\dropfiguresfalse
\ifdropfigures
\renewcommand{\includegraphics}[2][]{%
}
\fi

\usepackage{newfloat}
\usepackage{tikz}
\newcommand{\framedbox}[2][0.96\textwidth]{
 \centering
 \tikzstyle{mybox} = [draw=black,line width=1.2pt,inner sep=8pt]
 \begin{tikzpicture}
  \node [mybox] (box){%
   \begin{minipage}{#1}{#2}\end{minipage}
  };
 \end{tikzpicture}
}
\DeclareFloatingEnvironment[listname=Box,name=Box]{story}
\title{Language Agents as Digital Representatives\\
in Collective Decision-Making}

\author{%
~~\textbf{Daniel Jarrett}$^{*}$\\Google DeepMind
\\[-0.75ex]\\
\textbf{Michael Henry Tessler}\\Google DeepMind
\\[-0.75ex]\\
\textbf{Romuald Elie}\\Google DeepMind
\and
~~\textbf{Miruna P\^islar}$^{*}$\\Google DeepMind
\\[-0.75ex]\\
\textbf{Raphael K\"oster}\\Google DeepMind
\\[-0.75ex]\\
\textbf{Christopher Summerfield}\\Google DeepMind
\and
~~~\textbf{Michiel A. Bakker}~~~~~~\\~~~~Google DeepMind~~~~~~
\\[-0.75ex]\\
~~~\textbf{Jan Balaguer}~~~~~~\\~~~~Google DeepMind~~~~~~
\\[-0.75ex]\\
~~~\textbf{Andrea Tacchetti}~~~~~~\\~~~~Google DeepMind~~~~~~
}

\begin{document}

\maketitle
\allowdisplaybreaks


\begin{abstract}
\dayum{
Consider the process of collective decision-making, in which a group of individuals interactively select a preferred outcome from among a universe of alternatives.
In this context, ``representation'' is the activity of making an individual's preferences present in the process via participation by a proxy agent---i.e. their ``representative''.
To this end, learned models of human behavior have the potential to fill this role, with practical implications for multi-agent scenario studies and mechanism design.
In this work, we investigate the possibility of training \textit{language agents} to behave in the capacity of representatives of human agents, appropriately expressing the preferences of those individuals whom they stand for.
First, we formalize the setting of \textit{collective decision-making}---as the episodic process of interaction between a group of agents and a decision mechanism.
On this basis, we then formalize the problem of \textit{digital representation}---as the simulation of an agent's behavior to yield equivalent outcomes from the mechanism.
Finally, we conduct an empirical case study in the setting of \textit{consensus-finding} among diverse humans, and demonstrate the feasibility of fine-tuning large language models to act as digital representatives.
}
\end{abstract}
\section{Introduction}\label{sec:1}
\vspace{-0.25em}

\dayum{
Collective decision-making is a hallmark of intelligent behavior \cite{bose2017collective,mann2018collective},
and is ubiquitous in economic, social, and political spheres of society \cite{narahari2014game,leach2016freedom}.
Consider any process in which a group of individuals interactively select a preferred outcome from a universe of alternatives:
For instance, consider a diverse group of humans communicating in a mediated environment, seeking to arrive at a consensus opinion on a topic of debate \cite{bakker2022fine,tessler2023submit}.
In this context, ``representation'' is the activity of making an (otherwise absent) individual's preferences present in the process through participation by a proxy agent---i.e. their ``representative'' \cite{dovi2006political}.
To this end, learned models of human behavior have the potential to fill this role, with practical implications for scenario studies and mechanism design.
For instance, it can facilitate personalized and detailed simulations of collective interactions, allowing iterative refinement of mechanisms before real-world deployment.
The key question is: \textit{What makes a good representative}?
}

\dayum{
\textbf{Simulation for Representation}~
In this work, we explore the possibility of training \textit{language agents} to behave in the capacity of representatives of human agents, appropriately expressing the preferences of those individuals whom they stand for.
On the one hand,
this has natural connections to existing work in simulating human behavior,
such as
learning from demonstrations
\cite{le2016smooth,yue2018imitation,osa2018imitation,huyuk2021explaining},
generating synthetic data,
\cite{veselovsky2023generating,tang2023does,chan2021medkit, dosovitskiy2017carla},
and creating plausible simulacra of social agents
\cite{park2022social,aher2023using,harding2023ai,argyle2023out,horton2023large,park2023generative}.
On the other hand,
unlike such prior work,
our objective in simulating human behavior is specifically for ``representation''. This entails unique requirements:
(a) grounding in the context of \textit{collective interaction},
(b) attending to the granularity of \textit{individual fidelity}, and
(c) operating in the high-dimensional domain of \textit{language space}.
Specifically, in this work, we are exploring what representation means, how to measure representativity, and whether language agents can be trained to act as representatives on behalf of humans.
}

\dayum{
\textbf{Contributions}~
We make three key contributions in this investigation. First, we formalize the setting of \textit{collective decision-making} as the episodic process of interaction between a group of agents and a decision mechanism (Section \ref{sec:2}).
Building upon this, our second contribution is to characterize the problem of \textit{digital representation} as the simulation of an agent's behavior so as to yield equivalent outcomes when interacting with the mechanism (Section \ref{sec:3}).
Finally, we conduct an empirical case study in the setting of \textit{consensus-finding} in natural language among diverse humans, and demonstrate the feasibility of fine-tuning large language models to act as such digital representatives (Section \ref{sec:4}).
}
\section{Collective Decision-Making}\label{sec:2}

Consider a general setting for collective decision-making,
where interactions between a group of individuals are mediated by a \textit{decision mechanism},
yielding a process that produces a final outcome.
By way of preliminaries,
we first formalize the notion of a \textit{social choice function},
which is an abstract mapping from preferences to outcomes that is implemented by a mechanism:

\begin{redefinition}[restate=defsocial,name=Social Choice Function]\upshape\label{def:social}
%
Let $N:=\{1,\dots,n\}$ denote a set of \textit{participants},
who are engaged in the process of making a collective decision.
Denote with $\Omega$ the space of \textit{outcomes},
from which the participants are required to make a final selection $\omega\in\Omega$.
Denote with $\theta_{i}\in\Theta_{i}$ the \textit{type} characterizing each participant $i\in N$,
which captures their preferences over different outcomes.
Moreover, let $\theta:=(\theta_{i})_{i\in N}\in\Theta:=\Theta_{1}\times\dots\times\Theta_{n}$ indicate the \textit{type profile} of all participants.
Then a \textit{social choice function} $h$ is a mapping from the space of type profiles into the space of outcomes,
\begin{equation}
h:\Theta\rightarrow\Omega
\end{equation}
In line with related fields,
``participants'' may also be referred to as ``agents'' and ``players'',
and an ``outcome'' may be synonymous with an ``alternative'' and a ``collective decision''.
\EOD
\end{redefinition}

We operate in the standard setting for discrete-time Markov decision processes.
Let $x\in\mathcal{X}$ denote the \textit{state} variable:
While we keep notation simple,
$x$ may play the role of ``contexts'' and ``histories'' that capture all observations prior to the current time step.
Let $u_{i}\in\mathcal{U}_{i}$ denote the \textit{action} variable for $i\in N$,
and let $u:=(u_{i})_{i\in N}\in\mathcal{U}:=\mathcal{U}_{1}\times\dots\times\mathcal{U}_{n}$ indicate the \textit{action profile} for all participants.
In our setting,
$u$ will largely play the role of ``utterances'' in language space.
Denote with $T$ the finite length of each decision episode.
The outcome of each episode is precisely the terminal state $\omega:=x^{T}$.

\begin{redefinition}[restate=defmechanism,name=Decision Mechanism]\upshape\label{def:mechanism}
%
Each participant $i\in N$ is associated with a \textit{behavior} denoted with $\pi_{i}\in\Pie_{i}\subseteq\Delta(\mathcal{U}_{i})^{\mathcal{X}}$,
and let $\pi:=(\pi_{i})_{i\in N}\in\Pie:=\Pie_{1}\times\dots\times\Pie_{n}$ give the \textit{behavior profile} for all participants.
A \textit{decision mechanism} $\tau$ is a mapping from states and action profiles to next states,
\begin{equation}
\tau\in\Tau\subseteq\Delta(\mathcal{X})^{\mathcal{X}\times\mathcal{U}}
\end{equation}
and an \textit{outcome function} $f$ maps behavior profiles and mechanisms into distributions over outcomes.
In our Markov decision process setting,
the outcome function is simply the result of ``rolling out'' an episode of the interactive process between $\pi$ and $\tau$ to arrive at (a distribution over) terminal states,
\begin{equation}
f:\Pie\times\Tau\rightarrow\Delta(\Omega)
\end{equation}
A ``behavior'' may synonymously be referred to as a ``policy'' or ``strategy'',
and a ``mechanism'' may be referred to as a ``game'' or ``environment''.
Our setting is in general non-stationary: We may use superscripts $t\in\{1,\dots,T\}$ to indicate time steps, but omit them unless explicitly required.
\EOD
\end{redefinition}

Finally,
a mechanism $\tau$ is said to \textit{implement} a social choice function $h$ when $h(\theta)=f(\pi, \tau)$
for all $\pi\in\Pi$.
Both $\pi$, $\tau$ may depend on $\theta$, though not required.
Often,
mechanisms are designed with an \textit{optimization objective} in mind.
For instance,
let each participant be associated with a \textit{payoff function},
\begin{equation}
g_{i}:\Omega\times\Theta_{i}\rightarrow\mathbb{R}
\end{equation}
such that $g_{i}(\omega,\theta_{i})$ gives the payoff that participant $i\in N$ enjoys from the collective decision $\omega$.
As before, for convenience, let $g:=(g_{i})_{i\in N}$ and (with some abuse of notation) write $g:\Omega\times\Theta\rightarrow\mathbb{R}^{n}$.
Then a ``utilitarian'' social choice function is implemented as
$h(\theta)=f(\pi, \tau^{*})$,
by the mechanism:
\begin{equation}
\tau^{*}
\in
\underset{\tau\in\Tau}{\text{arg~max}}~
\mathbb{E}_{\omega\sim f(\pi,\tau)}
G(\omega,\theta)
\qquad
\text{where}
\qquad
G(\omega,\theta)
:=
\frac{1}{n}
\sum_{i=1}^{n}g_{i}(\omega,\theta_{i})
\label{eqn:utilitarian}
\end{equation}
Note that a ``payoff function'' is often interchangeable with a ``reward function'' or ``utility function''.
In this work, we use the scenario of \textit{consensus-finding} among individuals as our illustrative example.

\begin{figure}[H]
\centering
\vspace{-0.25em}
\includegraphics[width=0.99\columnwidth]{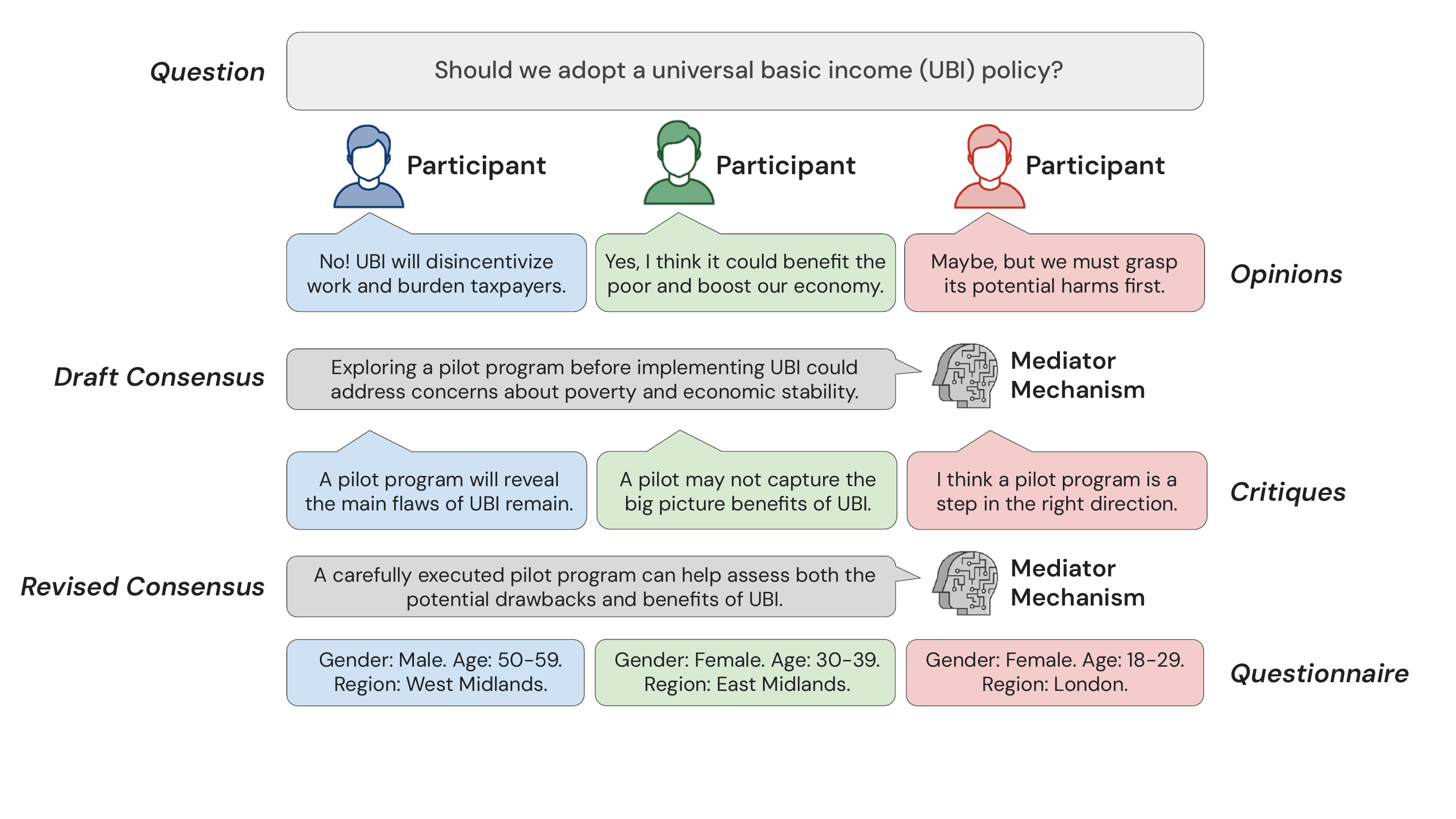}
\vspace{-2.25em}
\caption{\textit{Consensus-Finding}. Observe that this is an instance of a collective decision-making setting:}
\vspace{-1em}
\label{fig:diagram}
\end{figure}

\newpage
\textbf{Case Study (Consensus-Finding)}~
\dayum{
In this setting, a group of human participants shares their thoughts on a debated topic using natural language. Through interaction with a mediation mechanism, they aim to reach a consensus on the matter \cite{bakker2022fine,tessler2023submit}. Specifically, an episode begins with a set of $N$ participants observing a \textit{question} of interest, sampled from a corpus of questions. For instance, this may be: ``Should we adopt a universal basic income (UBI) policy?'' (see Figure \ref{fig:diagram}). Each participant expresses their \textit{opinion} in written text. Next, the mediator mechanism processes these opinions and outputs a \textit{draft consensus} statement. Each participant then provides their own \textit{critique} of the draft consensus in written text. Note that participants do not observe each other's opinions or critiques. Finally, the mediator mechanism processes these critiques to produce a \textit{revised consensus} statement. The episode ends with participants viewing the revised consensus, followed by an individual demographics \textit{questionnaire}.
}

\begin{tcolorbox}[rounded corners, left=8pt, right=7pt, top=14pt, bottom=7pt, boxsep=0pt, grow to left by=2pt, grow to right by=2pt, boxrule=0.75pt, colback=black!02, colframe=black!0]
\begin{itemize}[leftmargin=1em,labelsep=0.45em]
\itemsep-0.75pt
\vspace{-0.75em}
\item $N$ is the set of participants, typically 3--5 for each episode;
\item $\Omega=\mathcal{X}$ is the space of revised consensus statements;
\item $x^{1}\in\mathcal{X}$ is the \textit{question} of interest;
\item $u^{1}_{i}\in\mathcal{U}$ is the \textit{opinion} of participant $i\in N$ on the question;
\item $x^{2}\in\mathcal{X}$ is the \textit{draft consensus} statement taking those into account;
\item $u^{2}_{i}\in\mathcal{U}$ is the \textit{critique} of participant $i\in N$ on the draft consensus;
\item $x^{3}\in\mathcal{X}$ is the \textit{revised consensus} statement taking those into account;
\item $u^{3}_{i}\in\mathcal{U}$ is the dummy action that terminates the episode with a demographics questionnaire;
\item $\theta\in\Theta$ is the participants' preference profile over different consensus opinions;
\item $\pi_{i}(\cdot|x)$ is the utterance behavior of participant $i\in N$ in response to $x\in\mathcal{X}$; and
\item $\tau(\cdot|x,u)$ is the mediation mechanism's transition function at $x\in\mathcal{X}$, $u\in\mathcal{U}$.
\end{itemize}
\end{tcolorbox}
\vspace{0.5em}

\dayum{
\textbf{Remark (Mediator Mechanism)}~
Two comments deserve brief mention.
Firstly, for added context, the mediator mechanism $\tau$ is a 70B-parameter Chinchilla \cite{hoffmann2022training} model, fine-tuned as described in \cite{tessler2023submit} to best assist humans in collectively producing written consensus statements with maximal approval (viz. Equation \ref{eqn:utilitarian}). In particular, these statements are preferred more strongly than several high-quality baselines, including individual human opinions and consensus statements from human mediators.\footnote{The mediator mechanism was not trained to endorse a specific perspective or persuade others of any viewpoint; rather, it was trained purely to produce consensus statements based on the utterances provided by participants.}
Secondly, we want to emphasize that for our purposes, we treat the mechanism as a black-box transition function. Our goal is to investigate how \textit{digital representatives} can be trained to operate on behalf of humans in collective decision-making settings, with the consensus-finding scenario as our primary example. In this sense, the inner details of the mechanism are not of importance to us, save for recognizing that $\Tau$ (and $\Pie$, as we shall see) are families of pre-trained and fine-tuned large language models.
}
\section{Digital Representatives}\label{sec:3}

\dayum{
Suppose the true policy profile $\pi^{*}$\pix$\in$\pix$\Pie$, and we are interested in a model policy profile $\tilde{\pi}$ that approximates some or all of $\pi^{*}$ with digital representatives, in the context of decision mechanisms $\tau$\pix$\in$\pix$\Tau$. (For example, this might involve substituting a part of a single participant's policy, or multiple participants' policies, within $\pi^{*}$).
\textit{What defines the set of profiles $\tilde{\pi}$ that we can consider as ``equivalent'' to $\pi^{*}$?}
}

\subsection{Digital Clones}

One obvious choice is to require that $\pi^{*}(u|x)=\tilde{\pi}(u|x)$ for all $x\in\mathcal{X}$ and $u\in\mathcal{U}$, essentially seeking an identical clone of $\pi^{*}$.
This is fine as an objective for \textit{training} language agents to serve as digital representatives---in fact, likelihood-based training is exactly what we perform in Section~\ref{sec:4}.
However, we argue that matching conditionals alone is incomplete for \textit{evaluating} digital representatives, as it may be deemed both ``too strong'' and ``too weak''.
On one hand, it is ``too strong'' as $\tilde{\pi}$ may not need to clone the entirety of $\pi^{*}$---\textit{in the context of} $\tau$. Intuitively, suppose there were some decomposition,
\begin{equation}
\mathcal{U}=\mathcal{U}_{\para}\times\mathcal{U}_{\bot}
\end{equation}
such that $\mathcal{U}_{\para}$ somehow encapsulated the ``relevant'' dimensions of utterances, whereas $\mathcal{U}_{\bot}$ encapsulated the ``irrelevant'' dimensions.
It may be helpful to draw an analogy with the ``content'' vs. ``style'' distinction in the context of image classification \cite{wang2017effectiveness}, or the ``meaning'' vs. ``form'' distinction in the context of semantic clustering \cite{kuhn2023semantic}.
If this is true, we would only care about matching the conditional,
\begin{equation}
\pi(u_{\para}|x)
=
\textstyle\int_{\mathcal{U}_{\bot}}\pi(u|x)du_{\bot}
\end{equation}
That is, we would simply require that $
\pi^{*}(u_{\para}|x)
=
\tilde{\pi}(u_{\para}|x)
$, for $x\in\mathcal{X}$ and $u_{\para}\in\mathcal{U}_{\para}$.
\dayum{%
If we had access to such a decomposition, the matter would be settled.
But such a factoring is seldom apparent.

On the other hand, defining representativity purely on the basis of conditionals is also ``too weak'', as the policies in $\tilde{\pi}$ are ultimately unrolled in one or more transitions by interacting through the mechanism. Any discrepancy in conditionals may lead to compounding errors through such interactions.
}

\subsection{Digital Representatives}

A complete measure of representativity should also consider the dynamics of interaction.
Here, we draw connections with value-aware learning \cite{farahmand2017value,farahmand2018iterative} and value equivalence \cite{grimm2020value,grimm2021proper,grimm2022approximate,arumugam2022deciding} in model-based reinforcement learning. This allows us to define representation in a way that addresses both concerns at once.
First, some more notation:
Given any $\pi\in\Pie$, $\tau\in\Tau$,
the vector of \textit{expected payoffs} to participants at state $x$ taking action $u$ at time $t$ is given by the ``value function''
$Q_{\pi,\tau}^{t}:\mathcal{X}\times\mathcal{U}\rightarrow\mathbb{R}^{n}$:
\begin{equation}
Q_{\pi,\tau}^{t}(x,u)
=
\mathbb{E}_{x^{T}\sim f(\pi,\tau)}\big[g(x^{T},\theta)|x^{t}=x,u^{t}=u\big]
\end{equation}
Moreover, define the Bellman operator $\mathbb{B}_{\pi,\tau}$ over the space of functions
$Q:\mathcal{X}\times\mathcal{U}\rightarrow\mathbb{R}^{n}$ such that
\begin{equation}
(\mathbb{B}_{\pi,\tau}Q)(x,u)
:=
\mathbb{E}_{x'\sim\tau(\cdot|x,u)}
\mathbb{E}_{u'\sim\pi(\cdot|x')}
Q(x',u')
\label{eqn:bellman}
\end{equation}
\dayum{%
Then the sequence of functions \smash{$Q_{\pi,\tau}^{1:T}$} is the unique solution to backward recursions
\smash{$Q^{t}=\mathbb{B}_{\pi,\tau}^{t}Q^{t+1}$} for $t\in\{1,\dots,T-1\}$,
and $Q^{T}(x,u):=g(x,\theta)$.
Now we examine three notions of equivalence:
}

\begin{redefinition}[restate=defequivalence,name=Representational Equivalence]\upshape\label{def:equivalence}
Fix $\Pie$ and $\Tau$. Define first the set of policy profiles $\pi\in\Pie$ for which actions profiles $u\sim\pi(\cdot|x)$ are equal in distribution to $u\sim\pi^{*}(\cdot|x)$ given any state:
\begin{equation}
\Pie(\pi^{*})
:=
\{
\pi\in\Pie
:
\pi^{*}(\cdot|x)=\pi(\cdot|x)
~~
\forall
~
x\in\mathcal{X}
\}
\label{eqn:conditionals}
\end{equation}
Instead of matching \textit{conditionals} in isolation,
we may focus on \textit{transitions}:
Define the set of policy profiles $\pi\in\Pie$ whose operators $\mathbb{B}_{\pi,\tau}$ have equal effect to $\mathbb{B}_{\pi^{*},\tau}$ on any function $Q\in\mathcal{Q}\subseteq(\mathbb{R}^{n})^{\mathcal{X}\times\mathcal{U}}$:
\begin{equation}
\Pie(\pi^{*},\Tau,\mathcal{Q})
:=
\{
\pi\in\Pie
:
\mathbb{B}_{\pi^{*},\tau}Q
=
\mathbb{B}_{\pi,\tau}Q
~~~~
\forall
~
\tau\in\Tau
~
\text{and}
~
Q\in\mathcal{Q}
\}
\label{eqn:transitions}
\end{equation}
Finally, we may assess equivalence based on payoffs of \textit{trajectories}:
Define the set of policy profiles whose operators \smash{$\mathbb{B}_{\pi,\tau}^{1:T}$} have identical effect to \smash{$\mathbb{B}_{\pi^{*},\tau}^{1:T}$} when all applied onto any $Q^{T}\in\mathcal{Q}\subseteq(\mathbb{R}^{n})^{\mathcal{X}\times\mathcal{U}}$:
\begin{equation}
\Pie^{T}(\pi^{*},\Tau,\mathcal{Q})
:=
\{
\pi\in\Pie
:
\mathbb{B}_{\pi^{*},\tau}^{1}
\hspace{-8pt}\raisebox{1pt}{\scalebox{0.8}{$\circ\dots\circ$}}\hspace{1pt}
\mathbb{B}_{\pi^{*},\tau}^{T}Q^{T}
=
\mathbb{B}_{\pi,\tau}^{1}
\hspace{-7pt}\raisebox{1pt}{\scalebox{0.8}{$\circ\dots\circ$}}\hspace{1pt}
\mathbb{B}_{\pi,\tau}^{T}Q^{T}
~
\forall
~
\tau\in\Tau,
Q^{T}\in\mathcal{Q}
\}
\label{eqn:trajectories}
\end{equation}
Since repeated application of a Bellman operator $T$ times turns any function $Q^{T}$ into a value function for the beginning of an episode, this condition effectively asks for equality in expected payoffs.
\EOD
\end{redefinition}

Expression \ref{eqn:conditionals} is simply the singleton class of digital clones. However, Expressions \ref{eqn:transitions} and \ref{eqn:trajectories} both capture a notion of \textit{invariance} in $\Pie$ in the sense that they define potentially larger classes of policy profiles. Moreover, they are both defined on the basis of interaction with the mechanism. It turns out that Expression \ref{eqn:transitions} still demands too much, and \ref{eqn:trajectories} provides the better definition for ``representation'':

\begin{reproposition}[restate=thmequivalence,name=Representational Equivalence]\upshape\label{thm:equivalence}
Fix $\Pie$ and $\Tau$, and let $\mathcal{Q}$ be closed under Bellman updates. Consider the equivalence classes of policy profiles induced by $\pi^{*}$ from Definition \ref{def:equivalence}. We~have
\begin{equation}
\Pie(\pi^{*})
\subseteq
\Pie(\pi^{*},\Tau,\mathcal{Q})
\subseteq
\Pie^{T}(\pi^{*},\Tau,\mathcal{Q})
\label{eqn:subseteq}
\end{equation}
In particular, consider the maximal space of functions
$\mathcal{Q}=(\mathbb{R}^{n})^{\mathcal{X}\times\mathcal{U}}$.
If the space of mechanisms is also maximal,
that is $\Tau=\Delta(\mathcal{X})^{\mathcal{X}\times\mathcal{U}}$, then we have that the first subset relation is an equality, that is
\begin{equation}
\Pie(\pi^{*})
=
\Pie(\pi^{*},\Tau,\mathcal{Q})
\subseteq
\Pie^{T}(\pi^{*},\Tau,\mathcal{Q})
\label{eqn:equality}
\end{equation}
Let action profiles be decomposable as $\mathcal{U}=\mathcal{U}_{\para}\times\mathcal{U}_{\bot}$ with $\text{card}(\mathcal{U}_{\bot})$\pix$>$\pix$1$,
and the interaction between the mechanism and policies be such that values $Q_{\pi,\tau}^{t}(x,u)=Q_{\pi,\tau}^{t}(x,(u_{\para},u_{\bot}))=Q_{\pi,\tau}^{t}(x,u_{\para})$ for all $t<T$, $x\in\mathcal{X}$, $u\in\mathcal{U}$, $\pi\in\Pie$, and \smash{$\tau\in\Tau\subseteq\Delta(\mathcal{X})^{\mathcal{X}\times\mathcal{U}}$}. Then the second subset relation~is~proper,
\begin{equation}
\Pie(\pi^{*})
\subseteq
\Pie(\pi^{*},\Tau,\mathcal{Q})
\subset
\Pie^{T}(\pi^{*},\Tau,\mathcal{Q})
\label{eqn:subset}
\end{equation}
\end{reproposition}

\vspace{-0.5em}
\textit{Proof}. Appendix \ref{app:a}. \QED
\vspace{0.25em}

\dayum{
If the class of mechanisms is maximal, then the transition-based equivalence class $\Pie(\pi^{*},\Tau,\mathcal{Q})$ is reduced to a singleton, whereas it is still possible for the trajectory-based $\Pie^{T}(\pi^{*},\Tau,\mathcal{Q})$ to be larger (viz. Expression~\ref{eqn:equality}).
More importantly, if the class of mechanisms is such that expected payoffs are invariant to some $\mathcal{U}_{\bot}\subseteq\mathcal{U}$, then the equivalence class $\Pie^{T}(\pi^{*},\Tau,\mathcal{Q})$ is \textit{strictly} the largest.
The intuition is that $\tilde{\pi}\in\Pie^{T}(\pi^{*},\Tau,\mathcal{Q})$ are free to define arbitrary behavior within $\mathcal{U}_{\bot}$ without affecting expected payoffs of final outcomes.
As an example, suppose the class of mechanisms $\Tau\subset\Delta(\mathcal{X})^{\mathcal{X}\times\mathcal{U}}$ is itself invariant to some $\mathcal{U}_{\bot}\subseteq\mathcal{U}$
(that is, $\tau(\cdot|x,u)=\tau(\cdot|x,(u_{\para},u_{\bot}))=\tau(\cdot|x,u_{\para})$ for all $x\in\mathcal{X}$, $u\in\mathcal{U}$, $\tau\in\Tau$).
It is easy to see---by induction from \smash{$Q_{\pi,\tau}^{T}(x,u):=g(x,\theta)$}---that the invariance assumption on value functions required for Expression \ref{eqn:subset} holds (but not necessarily for Expression \ref{eqn:equality}).
}

\vspace{0.25em}
\textbf{Value Equivalence}~
In summary, this motivates a simple estimate of ``representativity'' that captures how much a given model profile $\tilde{\pi}$ deviates from the true profile $\pi^{*}$ through their expected payoffs:
\vspace{0.25em}
\begin{equation}
\textit{Representativity}(\pi^{*},\tilde{\pi})
:=
\hspace{-5pt}
\max_{\tau\in\Tau,Q^{T}\in\mathcal{Q}}
\mathcal{L}
\big(
\mathbb{E}_{\omega\sim f(\pi^{*},\tau)}Q^{T}(\omega)
,
\mathbb{E}_{\omega\sim f(\tilde{\pi},\tau)}Q^{T}(\omega)
\big)
\label{eqn:representativity}
\end{equation}
where $\mathcal{L}$ is some measure of discrepancy between its vector arguments. In other words, $\tilde{\pi}$ is a ``good'' representative of $\pi^{*}$ if behaviors elicited from each profile are such that, when aggregated through mechanisms $\tau$\pix$\in$\pix$\Tau$, they yield outcomes that (in expectation) are equivalent as measured by $Q^{T}$\pix$\in$\pix$\mathcal{Q}$.\footnote{\dayum{Note that this is reminiscent of how environment models are judged based on value equivalence \cite{farahmand2017value,farahmand2018iterative,grimm2020value,grimm2021proper,grimm2022approximate,arumugam2022deciding}. However, instead of defining equivalent environments based on a reference class of policies for reinforcement learning, here we are defining equivalence classes of multi-agent policy profiles based on a reference class of mechanisms.}\vspace{-1.25em}}
\section{Case Study: Consensus-Finding}\label{sec:4}

In the context of consensus-finding (viz. Figure \ref{fig:diagram}) as collective decision-making, we now turn to the empirical question:
\textit{Is it feasible to train language agents to act as digital representatives of humans?}
Indeed, large language models are often pre-trained on datasets rich in diverse preferences, and
recent work has shown that language agents can generate plausible behavior in interactive settings \cite{park2023generative}, and
to simulate subpopulations of humans in studies in
economics \cite{horton2023large},
psychology \cite{aher2023using}, and
social science \cite{argyle2023out}.
However, while socio-demographically prompted models can capture the sentiment of general subpopulations of humans \cite{santurkar2023whose,simmons2022moral}, they are not as reliable on more granular levels \cite{beck2023not,harding2023ai}. Is it possible to fine-tune language models on a personalized level to represent \textit{individuals} in a group?

\subsection{Experiment Setup}

In this work, we train and evaluate digital representatives using a consensus-finding dataset from \cite{tessler2023submit}. For completeness, we briefly recall the input questions and data collection process \cite{tessler2023submit}. Then we give an overview of the final dataset and the training process for digital representatives in our experiment.

\dayum{
\textbf{Questions}~
The corpus of input questions concern reasonably divisive political issues relevant in the United Kingdom. Questions were generated using a pre-trained 70B-parameter Chinchilla \cite{hoffmann2022training} language model. First, 175 ``seed'' questions on contemporary topics of debate were written by hand. Random samples of 10 seed questions at a time were used to prompt the model to generate final questions, for a total of 10,000 unique questions. Questions likely to prompt offensive commentary as well as those that were nonsensical or uncontentious were removed by hand, leaving a corpus of 5,645 questions. See \cite{tessler2023submit} for more detail.
}

\dayum{
\textbf{Data Collection}~
In each episode of data collection, a group of 3--5 participants based in the United Kingdom participated in an implementation of the protocol described in Figure \ref{fig:diagram}. To begin, each participant $\pi_{i}^{*}$ viewed a question $x^{1}$ sampled from the corpus. Next, they each wrote an opinion $u_{i}^{1}$ by typing free text into an input box in a custom web application. Then, the opinions of the group were included in a prompt provided to the mediator mechanism $\tau$ to generate a draft consensus statement $x^{2}$. Each participant then wrote a critique $u_{i}^{2}$ by typing free text into another input box. Finally, these were all included in another prompt provided to the mediator mechanism to generate a revised consensus statement $x^{3}$, after which participants answered a short demographic questionnaire $u_{i}^{3}$. Separately, before writing their initial opinion, participants were asked to provide a ``position'' score indicating their agreement on the declarative form of the question (so a question of the form ``Should we [...]'' will have declarative form ``We should [...]''), to allow measuring baseline disagreement among the group. For an experimental session, each set of participants were engaged in three different episodes (i.e., corresponding to three different questions), and an average session took 60 minutes.\footnote{Full details of the data collection procedure were first reviewed by an independent ethics review committee. No personally identifiable data was collected, and no offensive content was shown to participants. All participants provided informed consent before engaging and were compensated above the regional living wage.} See \cite{tessler2023submit} for more detail.
}

\dayum{
\textbf{Experiment Dataset}~
The final dataset we use for experiments is the result of 1,662 episodes of data collection,
corresponding to a total of 2,290 participants and 6,872 written opinions and critiques.
For each episode, the dataset includes the text of each question, written opinions, draft consensus statement, written critiques, and revised consensus statement.
For each participant in an episode, the metadata recorded include the other questions that the participant had engaged in from a past debate, as well as the written opinions and critiques of that participant on those other questions.
The questionnaire each participant completed covers a range of demographic information, including gender identity, age (in decades), ethnicity, region of domicile, education, political and religious affiliation, immigration status, and approximate household income.
For each question, the initial ``position'' score of each participant was also recorded.
We split this final dataset into training and validation sets on the basis of \textit{both} data collection episodes and participant identity.
This ensured that each participant and each question only appears in either the training set or the validation set, but not both.
This produced a training set with 1,332 episodes of data collection, corresponding to 1,828 participants and 5,484 instances of written opinions and written critiques, and a validation set with 330 episodes of data collection, corresponding to 462 participants and 1,386 instances of written opinions and critiques.
}

\begin{table}[htbp]\small
\newcolumntype{A}{>{\arraybackslash}m{ 2.7cm}}
\newcolumntype{B}{>{\arraybackslash}m{13.0cm}}
\caption{\textit{Example Critique}. Critiques sampled from a vanilla/fine-tuned digital representative (``DR'').}
\label{tab:example}
\begin{center}
\begin{adjustbox}{max width=1.005\textwidth}
\begin{tabular}{A|B}
\toprule
\textbf{Ground Truth}
&
I agree that some sort of tax should be mandatory but as I said earlier if its too severe these wealthy people wont [\textbf{sic}] stay on these islands and take their wealth elsewhere, its [\textbf{sic}] one thing having wealth but it has to attractive for these people.
\\
\midrule
\textbf{Vanilla DR 1B}
&
It is really quite unclear why the government wants to make this change. It does not make sense to make such a drastic change to the wealth tax. The logic of the government is that by introducing such a tax they will be able to raise £45 billion in revenues. The same money will be used to build infrastruct-[\textbf{...}]
\\
\midrule
\textbf{Fine-Tuned DR 1B}
&
I agree with this opinion and think it captures the nuances of the issue well.
\\
\midrule
\textbf{Vanilla DR 30B}
&
It is a little unclear to me why we should have this tax at all. I understand the problem of inequality, but I'm not sure that a flat tax would help at all. We could be doing more good by helping people on lower incomes to find jobs.
\\
\midrule
\textbf{Fine-Tuned DR 30B}
&
I agree with this. However the tax should not be so severe as to make people leave the country and therefore the gap would get wider.
\\
\bottomrule
\end{tabular}
\end{adjustbox}
\end{center}
\vspace{-1em}
\end{table}

\textbf{Digital Representatives}~
In our experiments, we focus on learning digital representatives of human participants for the \textit{critique} step ($u^{2}$).
Let $\hat{\pi}_{i}$ be a model trained to produce a written critique on the basis of a question, participant $i$'s opinion, a draft consensus, and (optionally) any additional information about that participant.
Then the digital representative $\tilde{\pi}_{i}$ for that participant is defined as:
\begin{equation}
\tilde{\pi}_{i}(\cdot|x^{t})
:=
\begin{cases}
\hat{\pi}_{i}\hspace{1.5pt}(\cdot|x^{t})&\text{if}~~t=2\\
\pi^{*}_{i}(\cdot|x^{t})&\text{otherwise}
\end{cases}
\end{equation}
Our training pipeline follows a standard supervised learning procedure for fine-tuning large language models to learn from demonstrations.
We train 1B- and 30B-parameter adaptations of Chinchilla \cite{hoffmann2022training} models as digital representatives $\hat{\pi}_{i}$, using the dataset to estimate and maximize the training objective:
\begin{equation}
\mathcal{J}_{\text{training}}
(\phi)
:=
\mathbb{E}_{\substack{
\pi_{i}^{*},\tilde{\pi}_{i}\sim\mathcal{D}_{\text{training}}\vspace{-5pt}\\
\vdots\\
u_{i}^{2}\sim\pi_{i}^{*}(\cdot|x^{2})
}}
\big[
\log\tilde{\pi}_{i}(u_{i}^{2}|x^{2})
\big]
\end{equation}
where $\phi$ denotes the trainable parameters of the model,
and the vertical ellipsis indicates the sampling of intermediate states and actions omitted for brevity.
We fine-tune all model parameters using a constant learning rate of $2 \cdot 10^{-6}$, a batch size of 128, and the AdamW optimizer \cite{adamw}. 
During training, we evaluate the model's performance by calculating the perplexity on the validation set. This metric is employed for early stopping to prevent overfitting. Following this procedure, we select the best checkpoints after 150 training steps for the 1B model and after 10 steps for the 30B model.
%

\subsection{Results: Critique Evaluation}

First, we evaluate digital representatives at the level of individual critiques. Our first performance measure is the log-likelihood of ground-truth critiques in the validation set, under the learned model:
\begin{equation}
\mathcal{J}_{\text{likelihood}}^{(\text{critique})}
(\phi)
:=
\mathbb{E}_{\substack{
\pi_{i}^{*},\tilde{\pi}_{i}\sim\mathcal{D}_{\text{validation}}\vspace{-5pt}\\
\vdots\\
u_{i}^{2}\sim\pi_{i}^{*}(\cdot|x^{2})
}}
\big[
\log\tilde{\pi}_{i}(u_{i}^{2}|x^{2})
\big]
\end{equation}
Another measure involves prompting a large version of PaLM 2 \cite{anil2023palm} (i.e. over 1--2 orders of magnitude larger than digital representatives) as an autorater, reporting the win-rate against a baseline critique:
\begin{equation}
\mathcal{J}_{\text{autorater}}^{(\text{critique})}
(\phi)
:=
\mathbb{E}_{\substack{
\pi_{i}^{*},\tilde{\pi}_{i}\sim\mathcal{D}_{\text{validation}}\vspace{-5pt}\\
\vdots\\
u_{i}^{2}\sim\tilde{\pi}_{i}(\cdot|x^{2})\\
v_{i}^{2}\sim\pi_{\text{baseline}}
}}
\big[
\textit{Autorater}(u_{i}^{2}\succ v_{i}^{2}|x^{2})
\big]
\end{equation}
where the golden baseline is the ground-truth critique (i.e. $\pi_{\text{baseline}}:=\pi^{*}$). See Table \ref{tab:example} for some critique samples. See Figure \ref{fig:critique-eval} for results on ablations and baseline (additional results in Appendix~\ref{app:bonus}).

\begin{figure}[htbp]
\centerline{
\includegraphics[width=0.41\columnwidth]{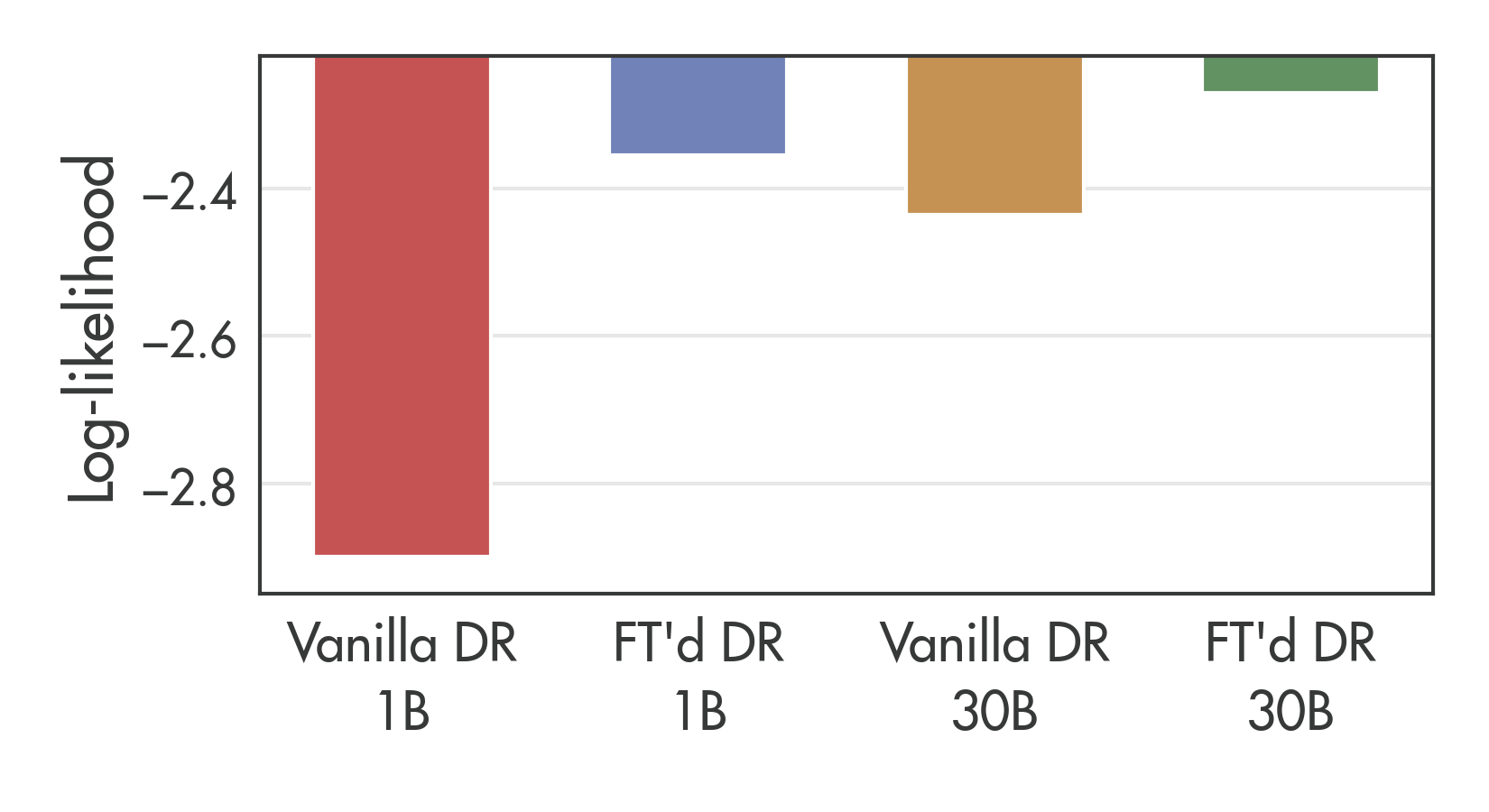}
\includegraphics[width=0.59\columnwidth]{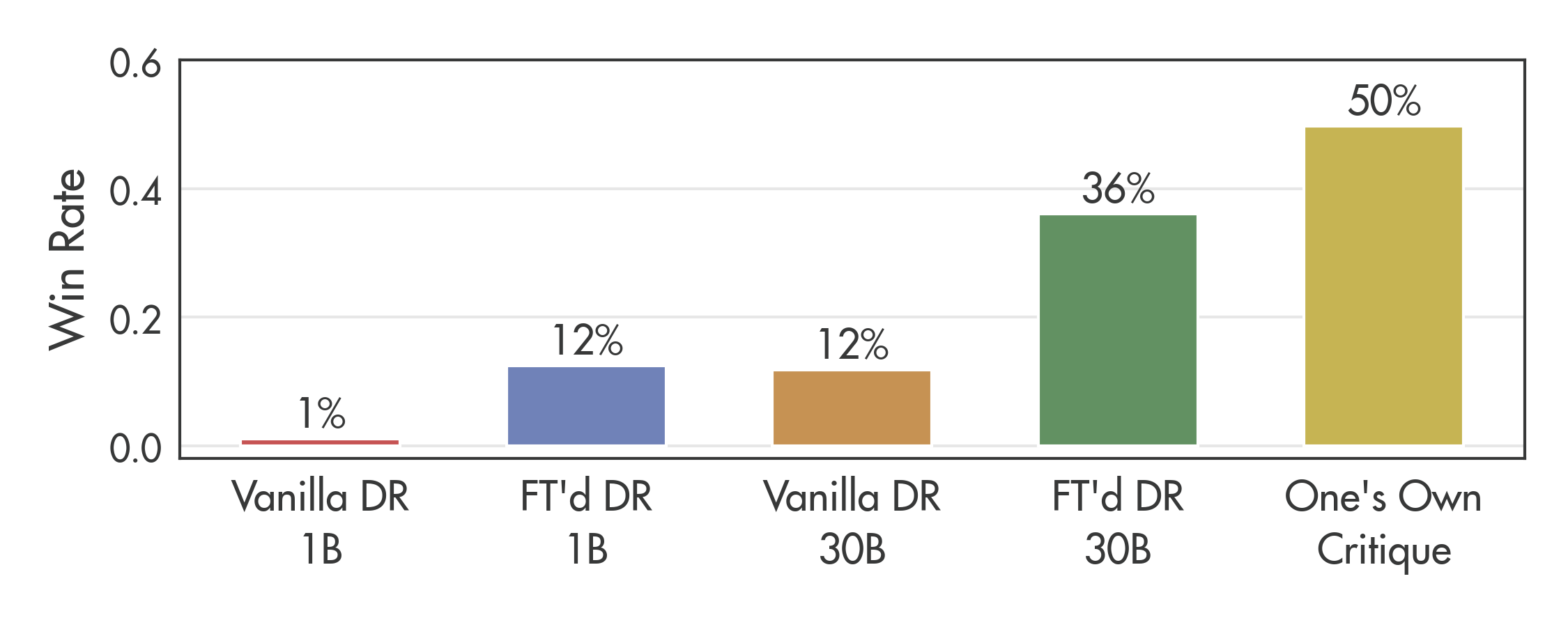}
}
\caption{\textit{Critique Evaluation}.
\textbf{Left}: Mean log-likelihood of ground-truth critiques from human participants (from the validation set), evaluated under models $\hat{\pi}_{i}$ with \textit{fine-tuned} (``FT'd'') or \textit{vanilla} digital representatives (``DRs'') with 1B or 30B parameters. The fine-tuned models consistently exhibit higher log-likelihoods compared to their vanilla counterparts, indicating a superior representation of participants' ground-truth critiques.
\textbf{Right}: The autorater's win-rate of a sampled critique (for different models $\hat{\pi}_{i}$ across the x-axis) against one's own ground-truth critique (i.e. the baseline). The golden bar represents $\pi^{*}$ against itself, serving as a reference for ceiling performance.
\textbf{Conclusion}: \textit{Fine-tuning and scale both improve the representativeness of DRs for held-out participants' critiques}.
}
\label{fig:critique-eval}
\end{figure}


\begin{figure}[htbp]
\centerline{
\includegraphics[width=0.41\columnwidth]{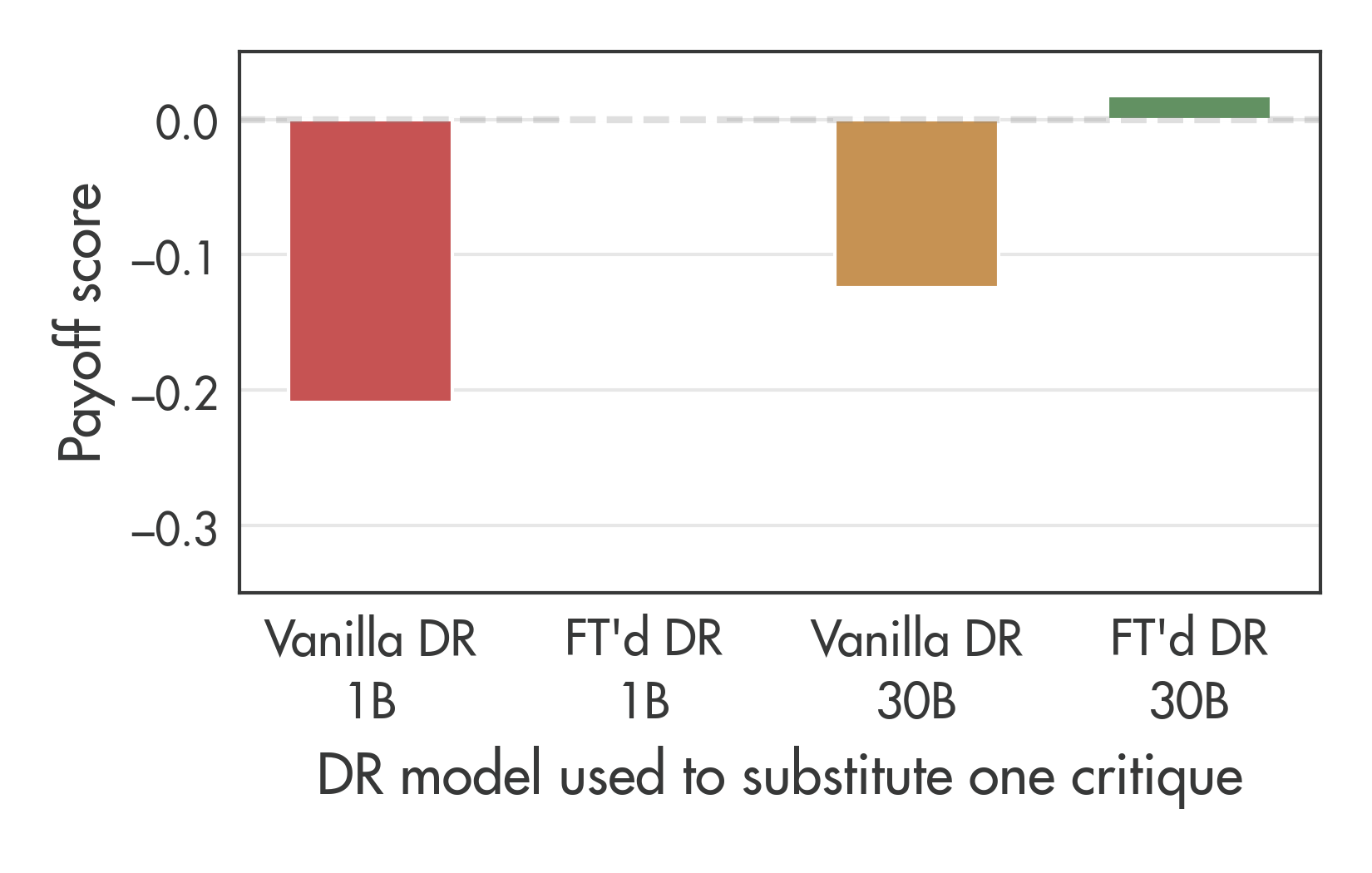}
\includegraphics[width=0.59\columnwidth]{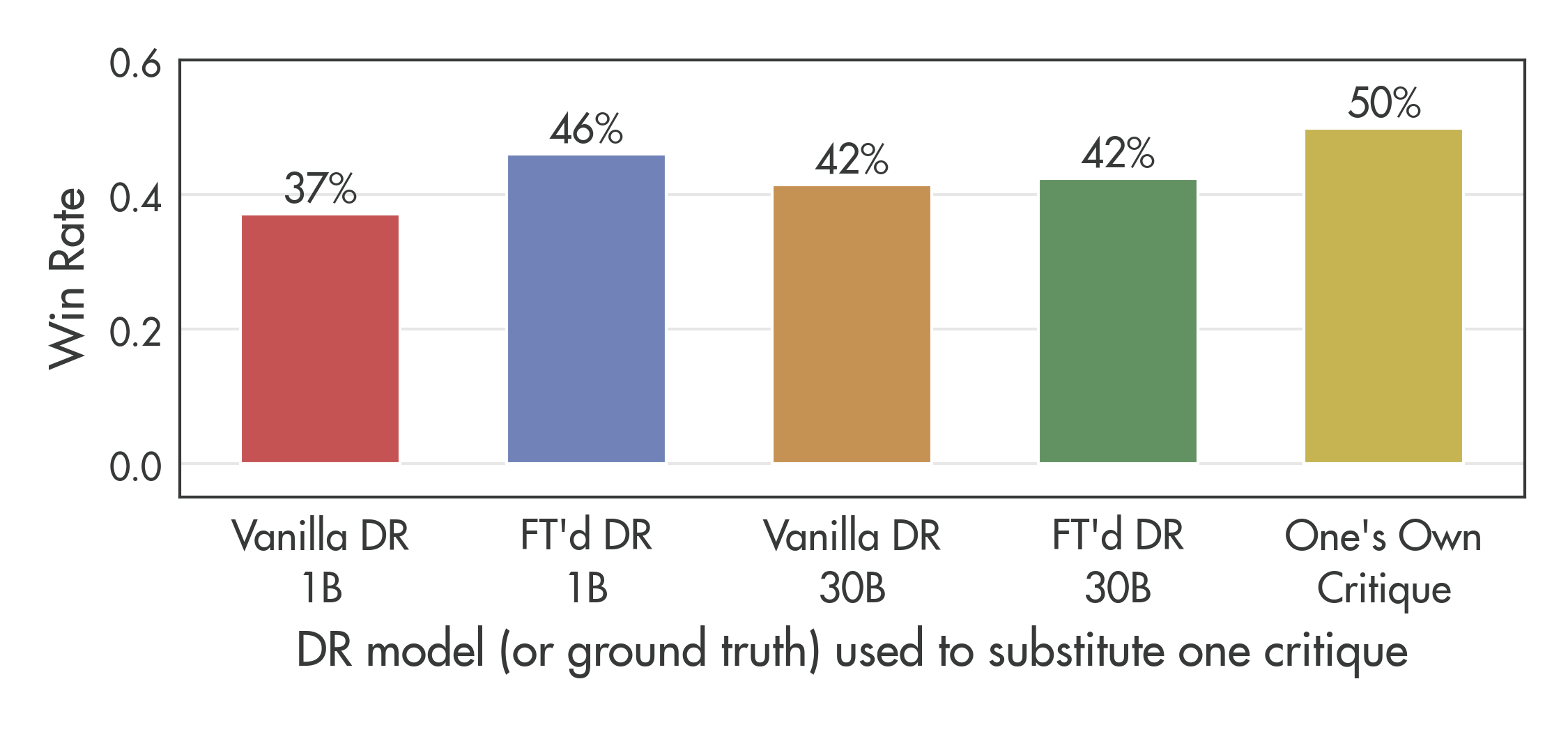}
}
\centerline{
\includegraphics[width=0.41\columnwidth]{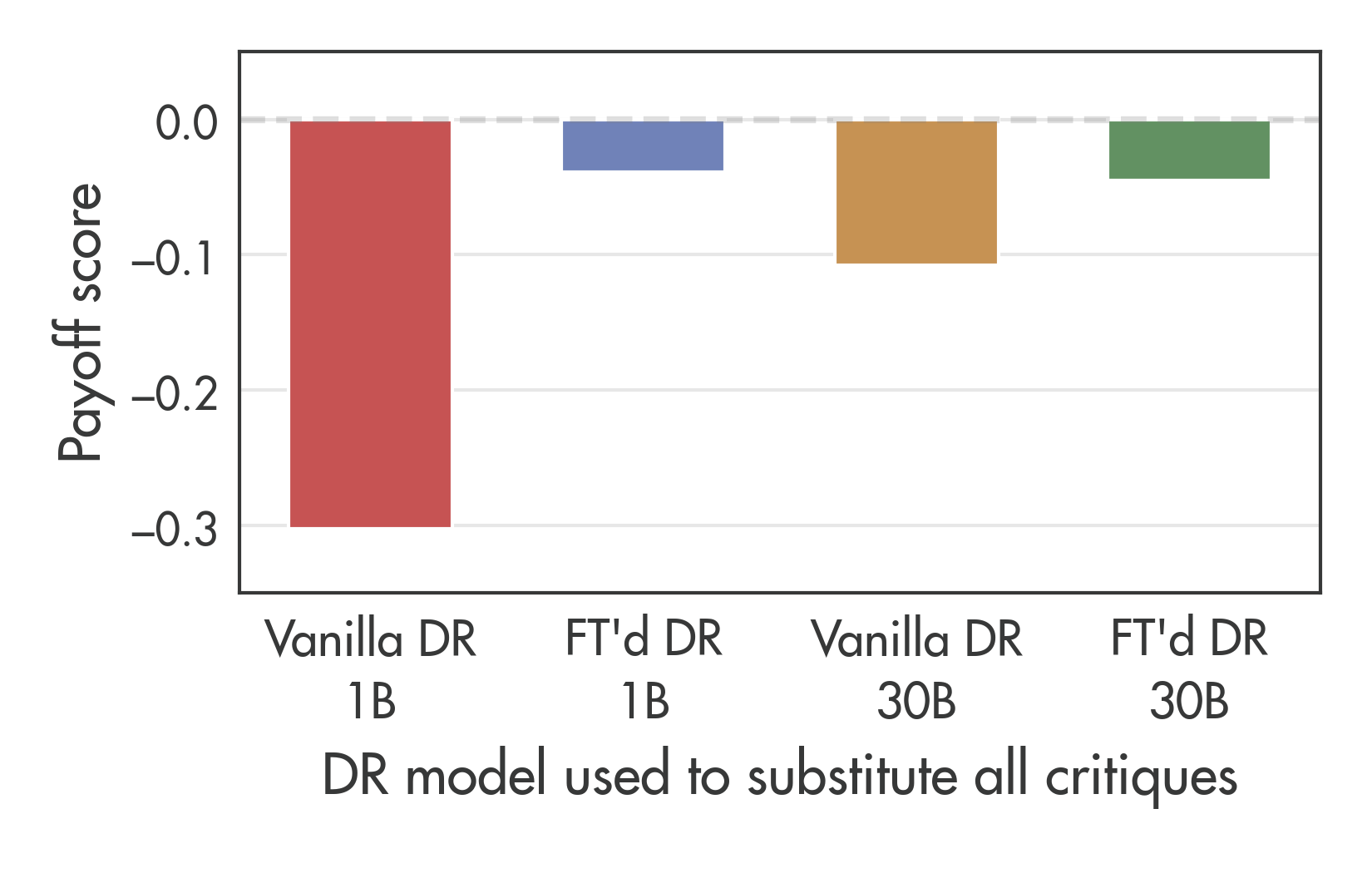}
\includegraphics[width=0.59\columnwidth]{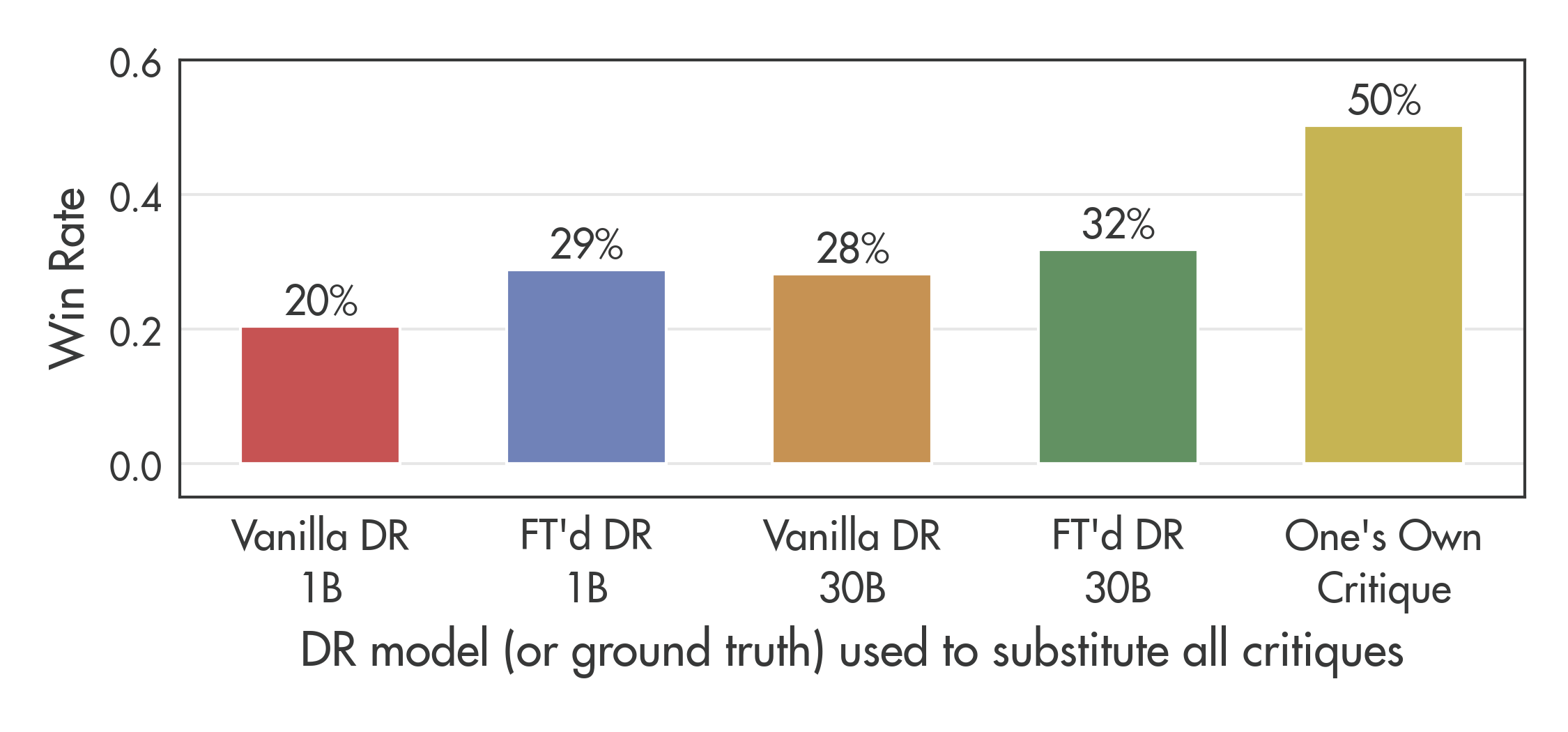}
}
\vspace{-0.1cm}
\caption{
\textit{Consensus Evaluation}. Either one participant's critique (\textbf{top}), or all participants' critiques (\textbf{bottom}) are substituted with critiques sampled from their respective digital representatives $\hat{\pi}_{i}$.
\textbf{Left}: Mean discrepancy in payoffs between the revised consensus generated by the mediator mechanism using participants' ground-truth critiques, versus using critiques sampled from DRs. Replacing either one or all ground truth critiques with vanilla DR samples results in significantly degraded payoffs, whereas fine-tuned DR samples yield payoffs with little or no degradation, indicating that those consensuses are more or less equivalent.
\textbf{Right}: The autorater's win-rate of a generated revised consensus (using critiques sampled from different models $\hat{\pi}_{i}$) against the revised consensus generated using ground-truth critiques (i.e. the baseline). The golden bar represents the baseline consensus against itself, serving as a reference for ceiling performance. Substituting a single critique (chosen at random) results in consensuses perceived as roughly equally similar across all DRs by the autorater (13\% difference between ceiling and Vanilla 1B DRs). This is likely a property of the mediation mechanism, which we empirically observe disregards outlier critiques. However, substituting the entire group's critiques significantly influences the revised consensus output (30\% difference between ceiling and Vanilla 1B DRs), with the 30B fine-tuned DRs having a notably higher win-rate here.
\textbf{Conclusion}: \textit{Using the notion of representativity motivated earlier, in both top and bottom panels we also observe that fine-tuning and scale both improve the representativity of DRs for held-out episodes}.
}
\label{fig:consensus-eval}
\vspace{-0.6cm}
\end{figure}

\subsection{Results: Consensus Evaluation}

Second, we evaluate digital representatives at the level of consensus outcomes. Our first performance measure is the discrepancy in expected payoff (cf. Equation \ref{eqn:representativity}) relative to some baseline consensus,
where payoffs are estimated using a payoff model accompanying the work in \cite{bakker2022fine,tessler2023submit}, which is a model based on a 1B-parameter Chinchilla \cite{hoffmann2022training} that outputs a scalar ``agreement'' score\footnote{The use of a payoff model serves as a proxy for human endorsement without replicating the entire study of \cite{bakker2022fine,tessler2023submit}.} for each participant:
\begin{equation}
\mathcal{J}_{\text{payoff}}^{(\text{outcome})}
(\phi)
:=
\mathbb{E}_{\substack{
\pi^{*},\tilde{\pi}\sim\mathcal{D}_{\text{validation}}\vspace{-5pt}\\
\vdots\\
u^{2}\sim\tilde{\pi}(\cdot|x^{2})\\
x^{3}\sim\tau(\cdot|x^{2},u^{2})\\
y^{3}\sim\tau_{\text{baseline}}
}}
\big[
\mathcal{L}
\big(
Q^{3}(x^{3}),Q^{3}(y^{3})
\big)
\big]
\end{equation}
Here, $\mathcal{L}$ computes the average element-wise difference between its two inputs, for participants who are substituted by their digital representatives.
We test two regimes of model policy profiles $\tilde{\pi}$:
One in which a single participant is substituted,
that is $
\tilde{\pi}_{-i}
:=
(\pi^{*}_{1},\dots,\tilde{\pi}_{i},\dots,\pi^{*}_{n})
$,
and one in which all of them are substituted,
that is $
\tilde{\pi}_{N}
:=
(\tilde{\pi}_{1},\dots,\tilde{\pi}_{n})
$.
As before, another measure involves an autorater:
\begin{equation}
\mathcal{J}_{\text{autorater}}^{(\text{outcome})}
(\phi)
:=
\mathbb{E}_{\substack{
\pi^{*},\tilde{\pi}\sim\mathcal{D}_{\text{validation}}\vspace{-5pt}\\
\vdots\\
u^{2}\sim\tilde{\pi}(\cdot|x^{2})\\
v^{2}\sim\pi^{*}(\cdot|x^{2})\\
x^{3}\sim\tau(\cdot|x^{2},u^{2})\\
y^{3}\sim\tau_{\text{baseline}}
}}
\big[
\textit{Autorater}(x^{3}\succ y^{3}|x^{2},v^{2})
\big]
\end{equation}
where we report the win-rate against a baseline consensus (and the golden baseline is the ground-truth consensus). See Figure \ref{fig:consensus-eval} for results on ablations and baselines (and further results in Appendix~\ref{app:bonus}).

\section{Discussion}\label{sec:5}
We fine-tune language agents to act as digital representatives in the context of collective decision-making, offering practical implications for scenario studies and mechanism design. In the sequel, we provide an overview of related work and conclude with a discussion on limitations and future work.

\dayum{
\textbf{Related Work}~
Our work in ``simulation for representation'' straddles two domains of research: in (1) how human behavior is simulated with models, and (2) how human behavior is represented by models.
}

\muline{\textit{Simulation}}:~ First, \textit{imitation learning} deals with training an artificial agent to mimic a demonstrator \cite{le2016smooth,yue2018imitation,osa2018imitation,huyuk2021explaining},
to match some notion of performance
\cite{syed2010reduction,abbeel2004apprenticeship,neu2007apprenticeship,babes2011apprenticeship}, or
to recover some underlying motive for their behavior
\cite{ziebart2008maximum,finn2016guided,ho2016generative,jarrett2021inverse}.
Multiple artificial agents have also been cloned \cite{zhan2018generative,codevilla2019exploring}, such as for
simple control tasks \cite{song2018multi},
bandit environments \cite{shih2022conditional}, and
driving simulation \cite{bhattacharyya2018multi}.
This contrasts with our high-dimensional language domain, and with our focus on ``representativity'' in collective decision-making.
Second, \textit{synthetic data generation} deals with sampling from learned distributions to approximate real phenomena, such as in sequential data
\cite{lin2019generating,xu2020cot,jarrett2021time},
medical environments
\cite{chan2021medkit,rcgan},
driving simulation
\cite{dosovitskiy2017carla},
as well as in language space
for social science
\cite{veselovsky2023generating},
privacy preservation,
\cite{kurakin2023harnessing},
and to augment
text mining
\cite{tang2023does}.
However, this contrasts with our emphasis on learning models to act in an interactive context.
Third, recent work has explored creating plausible simulacra of \textit{social agents} using language models, such as
in creating populated prototypes for social computing
\cite{park2022social},
in prompting models to simulate human sub-populations
\cite{argyle2023out}
and replicate human subject studies
\cite{aher2023using,harding2023ai},
in simulating economic agents
\cite{horton2023large},
and in creating believable social behavior in sandbox environments
\cite{park2023generative}.
In contrast, we fine-tune models on a personalized level to represent specific individuals within a group.

\dayum{
\muline{\textit{Representation}}:~
Since large language models are often pre-trained on human data with diverse perspectives,
a recent line of work has seeks to measure the preferences that pre-trained models \textit{inherently reflect} with prompting, such as
the representation of broad-based global demographic groups
\cite{durmus2023towards},
the alignment of their views with demographic categories in the United States
\cite{santurkar2023whose},
as well as to uncover the intrinsic ideological biases that pre-trained models exhibit
\cite{hartmann2023political}.
Secondly, a related strand of research systematically probes pre-trained models to \textit{actively simulate} biased perspectives, such as
by prompting with socio-demographic profiles
\cite{beck2023not}, as well as
by prompting with liberal or conservative profiles to sample text with corresponding moral biases
\cite{simmons2022moral}.
Some models have also been \textit{explicitly fine-tuned} to query the opinions and worldviews of demographic categories of people \cite{jiang2022communitylm,haller2023opiniongpt}.
In contrast, however, in this work we seek to simulate behavior at the granularity of individual representativity, specifically in the context of collective interaction.
Finally,
while we draw an analogy with value-aware
\cite{farahmand2017value,farahmand2018iterative}
and value equivalent learning
\cite{grimm2020value,grimm2021proper,grimm2022approximate,arumugam2022deciding},
our work has close connections with broader notions of representation \cite{dovi2006political}
in collective interaction
\cite{narahari2014game},
consensus-based decision-making
\cite{leach2016freedom},
and has potential implications for real and simulated deliberative democracy
\cite{dryzek2011toward,engelstad1989assignment,bessette1994mild,leike2023proposal}.
}

\dayum{
\textbf{Future Work}~
We began with the question: ``\textit{What makes a good representative?}''. Our argument contends that a representative should not only capture the conditional dynamics of behavior but also ensure that its interaction through a mechanism preserves the trajectory-wise dynamics of outcomes.
Several caveats are in order:
First, there is no escaping the fact that the fidelity of any notion of ``representative'' is most solidly grounded in human endorsement. While we use a black-box payoff model as proxy, future work would greatly benefit from human validation of digital representatives.
Second, in this work we focused on learning representatives for the critique-writing step only. Future work would benefit from naturally extending this to the opinion-writing step as well, yielding fully-simulated representatives.
%
Thirdly, although we defined and evaluated representatives based on an equivalence objective, in our experiments the models were trained on a standard likelihood-based fine-tuning objective. Future work could explore directly training for the equivalence objective.
}

\textbf{Broader Impact}~
It is important to emphasize that our  training of digital representatives is geared toward \textit{simulation}, not advocating for their deployment as substitutes for human accountability in decision-making. For instance, suppose we were interested in using digital representatives for the purpose of simulating outcomes to improve some downstream tasks. Then evaluation of \textit{those} tasks must be performed with real humans. This is crucial, since we do not train representatives on any notion of factuality or transparency. In fact, representatives are specifically trained to mimic their human inputs per se, thereby carrying over any biases from those corresponding humans. That said, this research was conducted with a focus on societal benefit. The ultimate goal is to facilitate granular simulations of collective interactions for policy design, allowing decision mechanisms to benefit from scalable and cost-effective iterative refinement before real-world deployment.


\clearpage
\bibliographystyle{unsrt}
\bibliography{
reference/reference,
reference/0-imitate,
reference/1-reinforce,
reference/2-entropy,
reference/3-information,
reference/4-constraints,
reference/5-risk,
reference/6-intrinsic,
reference/7-bounded,
reference/8-identification,
reference/9-interpret,
reference/a-miscellaneous,
reference/b-time,
reference/c-yoon
}


\clearpage
\appendix
\section{Proof of Proposition 1}\label{app:a}

\thmequivalence*

\textit{Proof}.
The first subset in Expression \ref{eqn:subseteq} is immediate from Definitions \ref{eqn:bellman} and \ref{eqn:transitions}.
The second subset can be seen as follows:
%
%
Fix $\tilde{\pi}\in\Pie(\pi^{*},\Tau,\mathcal{Q})$,
so for any $\tau\in\Tau$
and $Q\in\mathcal{Q}$
we have the following,
\begin{equation}
\mathbb{B}_{\pi^{*},\tau}Q
=
\mathbb{B}_{\tilde{\pi},\tau}Q
\end{equation}
But $\mathbb{B}_{\pi^{*},\tau}Q$ and $\mathbb{B}_{\tilde{\pi},\tau}Q$ are also in $\mathcal{Q}$ from the closure assumption, so we can repeatedly apply the Bellman operators $\mathbb{B}_{\pi^{*},\tau}$ and $\mathbb{B}_{\tilde{\pi},\tau}$ on the left and right hand sides for a total of $T$ times, therefore
\begin{equation}
\mathbb{B}_{\pi^{*},\tau}^{1}
\hspace{-8pt}\raisebox{1pt}{\scalebox{0.8}{$\circ\dots\circ$}}\hspace{1pt}
\mathbb{B}_{\pi^{*},\tau}^{T}Q
=
\mathbb{B}_{\tilde{\pi},\tau}^{1}
\hspace{-7pt}\raisebox{1pt}{\scalebox{0.8}{$\circ\dots\circ$}}\hspace{1pt}
\mathbb{B}_{\tilde{\pi},\tau}^{T}Q
\end{equation}
so $\tilde{\pi}\in\Pie^{T}(\pi^{*},\Tau,\mathcal{Q})$.
Next, the equality in Expression \ref{eqn:equality} is obvious.
%
%
Finally, 
%
%
the proper subset in Expression \ref{eqn:subset} can be shown by picking the following model policy $\tilde{\pi}$ as proof (note that $\tilde{\pi}\not\in\Pie(\pi^{*})$):
\begin{equation}
\tilde{\pi}((u_{\para},u_{\bot})|x)
:=
\mathbbl{1}_{\{u_{\bot}=\tilde{u}_{\bot}\}}
\pi^{*}(u_{\para}|x)
\end{equation}
First, note that the value function $Q_{\pi^{*},\tau}^{1:T}$ for the true policy $\pi^{*}$ profile satisfies the following recursion:
\begin{align}
Q_{\pi^{*},\tau}^{t}(x,(&u_{\para},u_{\bot}))
\\
&
=
(\mathbb{B}_{\pi^{*},\tau}^{t}Q_{\pi^{*},\tau}^{t+1})
(x,(u_{\para},u_{\bot}))
\\
&
=
\mathbb{E}_{x'\sim\tau(\cdot|x,(u_{\para},u_{\bot}))}
\mathbb{E}_{(u_{\para}',u_{\bot}')\sim\pi^{*}(\cdot|x')}
Q_{\pi^{*},\tau}^{t+1}(x',(u_{\para}',u_{\bot}'))
\\
&
=
\mathbb{E}_{x'\sim\tau(\cdot|x,(u_{\para},u_{\bot}))}
\textstyle\int_{\mathcal{U}_{\para}}
\textstyle\int_{\mathcal{U}_{\bot}}
\pi^{*}((u_{\para}',u_{\bot}')|x')
Q_{\pi^{*},\tau}^{t+1}(x',(u_{\para}',u_{\bot}'))
du_{\para}'du_{\bot}'
\\
&
=
\mathbb{E}_{x'\sim\tau(\cdot|x,(u_{\para},u_{\bot}))}
\textstyle\int_{\mathcal{U}_{\para}}
\textstyle\int_{\mathcal{U}_{\bot}}
\pi^{*}(u_{\para}'|x')
\pi^{*}(u_{\bot}'|x')
Q_{\pi^{*},\tau}^{t+1}(x',(u_{\para}',u_{\bot}'))
du_{\para}'du_{\bot}'
\\
&
=
\mathbb{E}_{x'\sim\tau(\cdot|x,(u_{\para},u_{\bot}))}
\textstyle\int_{\mathcal{U}_{\para}}
\pi^{*}(u_{\para}'|x')
Q_{\pi^{*},\tau}^{t+1}(x',u_{\para}')
du_{\para}'
\\
&
=
\mathbb{E}_{x'\sim\tau(\cdot|x,(u_{\para},u_{\bot}))}
\mathbb{E}_{u_{\para}'\sim\pi^{*}(\cdot|x')}
Q_{\pi^{*},\tau}^{t+1}(x',u_{\para}')
\end{align}
for $t\in\{1,\dots,T-1\}$, and $Q_{\pi^{*},\tau}^{T}(x,u)=g(x,\theta)$.
Likewise, the value function $Q_{\pi^{*},\tau}^{1:T}$ for the model policy profile $\tilde{\pi}$ satisfies the following recursion:
\begin{align}\textstyle
Q^{t}(x,(&u_{\para},u_{\bot}))
\\
&
=
(\mathbb{B}_{\tilde{\pi},\tau}^{t}Q^{t+1})
(x,(u_{\para},u_{\bot}))
\\
&
=
\mathbb{E}_{x'\sim\tau(\cdot|x,(u_{\para},u_{\bot}))}
\mathbb{E}_{(u_{\para}',u_{\bot}')\sim\tilde{\pi}(\cdot|x')}
Q^{t+1}(x',(u_{\para}',u_{\bot}'))
\\
&
=
\mathbb{E}_{x'\sim\tau(\cdot|x,(u_{\para},u_{\bot}))}
\textstyle\int_{\mathcal{U}_{\para}}
\textstyle\int_{\mathcal{U}_{\bot}}
\tilde{\pi}((u_{\para}',u_{\bot}')|x')
Q^{t+1}(x',(u_{\para}',u_{\bot}'))
du_{\para}'du_{\bot}'
\\
&
=
\mathbb{E}_{x'\sim\tau(\cdot|x,(u_{\para},u_{\bot}))}
\textstyle\int_{\mathcal{U}_{\para}}
\textstyle\int_{\mathcal{U}_{\bot}}
\mathbbl{1}_{\{u_{\bot}'=\tilde{u}_{\bot}\}}
\pi^{*}(u_{\para}'|x')
Q^{t+1}(x',(u_{\para}',u_{\bot}'))
du_{\para}'du_{\bot}'
\\
&
=
\mathbb{E}_{x'\sim\tau(\cdot|x,(u_{\para},u_{\bot}))}
\textstyle\int_{\mathcal{U}_{\para}}
\pi^{*}(u_{\para}'|x')
Q^{t+1}(x',(u_{\para}',\tilde{u}_{\bot}))
du_{\para}'
\\
&
=
\mathbb{E}_{x'\sim\tau(\cdot|x,(u_{\para},u_{\bot}))}
\mathbb{E}_{u_{\para}'\sim\pi^{*}(\cdot|x')}
Q^{t+1}(x',(u_{\para}',\tilde{u}_{\bot}))
\end{align}
for $t\in\{1,\dots,T-1\}$, and $Q^{T}(x,u)=g(x,\theta)$.
But $Q_{\pi^{*},\tau}^{1:T}$ is a solution to this recursion, and as solutions to backward recursions are unique, we have that $Q_{\tilde{\pi},\tau}^{1:T}=Q_{\pi^{*},\tau}^{1:T}$,
hence $\tilde{\pi}\in\Pie^{T}(\pi^{*},\Tau,\mathcal{Q})$.

Next, we show $\tilde{\pi}\not\in\Pie(\pi^{*},\Tau)$:
Pick $Q(x,(u_{\para},u_{\bot})):=\mathbbl{1}_{\{u_{\bot}\neq\tilde{u}_{\bot}\}}$. Then we have that
\begin{align}
(\mathbb{B}_{\pi^{*},\tau}Q)
(x,(u_{\para},u_{\bot}))
&
=
\mathbb{E}_{x'\sim\tau(\cdot|x,(u_{\para},u_{\bot}))}
\mathbb{E}_{(u_{\para}',u_{\bot}')\sim\pi^{*}(\cdot|x')}
Q(x',(u_{\para}',u_{\bot}'))
\\
&
=
\mathbb{E}_{x'\sim\tau(\cdot|x,(u_{\para},u_{\bot}))}
\mathbb{P}(u_{\bot}'\neq\tilde{u}_{\bot}|x',\pi^{*})
\\
&
=
\mathbb{P}(u_{\bot}'\neq\tilde{u}_{\bot}|x,(u_{\para},u_{\bot}),\pi^{*},\tau)
\end{align}

\vspace{-1.5em}
and likewise,
\vspace{-0.75em}

\begin{align}
(\mathbb{B}_{\tilde{\pi},\tau}Q)
(x,(u_{\para},u_{\bot}))
&
=
\mathbb{E}_{x'\sim\tau(\cdot|x,(u_{\para},u_{\bot}))}
\mathbb{E}_{u_{\para}'\sim\pi^{*}(\cdot|x')}
Q(x',(u_{\para}',\tilde{u}_{\bot}))
\\
&
=
\mathbb{E}_{x'\sim\tau(\cdot|x,(u_{\para},u_{\bot}))}
\mathbb{P}(\tilde{u}_{\bot}\neq\tilde{u}_{\bot})
\\
&
=
0
\end{align}
Suppose it were true that $\tilde{\pi}\in\Pie(\pi^{*},\Tau,\mathcal{Q})$, so that for all $Q$ we have that
$
(\mathbb{B}_{\pi^{*},\tau}Q)(x,(u_{\para},u_{\bot}))
=
(\mathbb{B}_{\tilde{\pi},\tau}Q)(x,(u_{\para},u_{\bot}))
$,
which means
$
\mathbb{P}(u_{\bot}'=\tilde{u}_{\bot}|x,(u_{\para},u_{\bot}),\pi^{*},\tau)
=
1
$.
But this is only possible if $\mathcal{U}_{\bot}$\pix$=$\pix$\{\tilde{u}_{\bot}\}$ hence $\text{card}(\mathcal{U}_{\bot})$\pix$=$\pix$1$, which is a contradiction with the premise that $\text{card}(\mathcal{U}_{\bot})$\pix$>$\pix$1$.
\QED
\section{Further Experiment Detail}\label{app:bonus}

\textbf{Variants of Digital Representatives}~ As a reminder, a digital representative $\hat{\pi}_{i}$ is a language model trained to produce a critique based on a question, participant $i$'s opinion, a draft consensus (all of which form the \textit{base} information), and optionally any supplemental information about participant $i$. The main text discusses various information sources within the task pipeline, which we outline in Table \ref{tab:prompt-info} along with text examples from the dataset. We experimented with different combinations of conditioning information along various axes of variation to create diverse datasets for fine-tuning. In Figure \ref{fig:dr-variants} we show a handful of variants of digital representatives when conditioning on various data in addition to the base information. Guided by the log-likelihood of the ground truth critiques (in the validation set), we empirically determined the most effective additional information, which included the participant $i$'s opinions and critiques on other debate questions discussed within the same data collection pipeline. An input prompt example from the dataset used to fine-tune the digital representatives as presented in the main text is illustrated in text box \ref{box:prompt-example}.

\vspace{-0.75em}
\begin{table}[H]\small
\newcolumntype{A}{>{\arraybackslash}m{ 0.6cm}}
\newcolumntype{B}{>{\arraybackslash}m{ 5cm}}
\newcolumntype{C}{>{\arraybackslash}m{10.2cm}}
\caption{\textit{Possible prompt information}. A list of sections (with examples) that may be included in the prompt for fine-tuning digital representatives.}
\label{tab:prompt-info}
\begin{center}
\begin{adjustbox}{max width=1.005\textwidth}
\begin{tabular}{A|B|C}
\toprule
\textbf{Alias}
&
\textbf{Description}
&
\textbf{Example section as part of the prompt}
\\
\toprule
\textbf{Base}
&
Current question under debate
&
Current question under debate: Should the government spend more on science and technology research?
\\
\midrule
\textbf{Base}
&
Opinion of participant $i$ about the question under debate
&
Opinion of the participant about the question under debate: I think its really important that the government spends more on Science and Technology. If you think about the problems we have as a society and as a planet - the answers lie with Science and Technology.
\\
\midrule
\textbf{Base}
&
Draft consensus statement
&
Group draft consensus statement (to be critiqued): The government should spend more on science and technology research as this would lead to better technology and medical discoveries to help the world.
\\
\midrule
\textbf{D}
&
Demographic profile of participant $i$ (based on questionnaire)
&
You represent a participant with the following demographic profile. Gender Identity: Male. Age (in decades): 50 to 59. Region: West Midlands. Ethnicity: White. Party voted for in the most recent General Election: Labour. Highest level of education completed: Higher or secondary or further education (A-levels, BTEC, etc.). Religious affiliation: No religion. Approximate household income: £20k to £40k per year. Immigration status in the UK: British Passport holder.
\\
\midrule
\textbf{O}
&
Participant $i$’s opinions about other questions (1 to 3)
&
Question 1 from a past debate: Should the government spend more on science and technology research?

Opinion of the participant about question 1: I believe that the government should be allocating more money for science and technology research.
\\
\midrule
\textbf{C}
&
Participant $i$’s critiques of other questions (1 to 3)
&
Question 1 from a past debate: Should the government spend more on science and technology research?

Critique of the participant about question 1: I totally agree. Science and Technology can help us out of our current problems and make life better for everyone. 
\\
\midrule
\textbf{P}
&
Participant $i$’s position score (as text) about other questions (1 to 3)
&
Question 1 from a past debate: Should the government spend more on science and technology research?

Position of the participant about question 1: agree.

Question 2 from a past debate: Should students be required to pass a literacy and numeracy test before graduating from primary school?

Position of the participant about question 2: strongly disagree.
\\
\bottomrule
\end{tabular}
\end{adjustbox}
\end{center}
\vspace{-1em}
\end{table}

\begin{figure}[H]
\centering
\includegraphics[width=0.96\columnwidth]{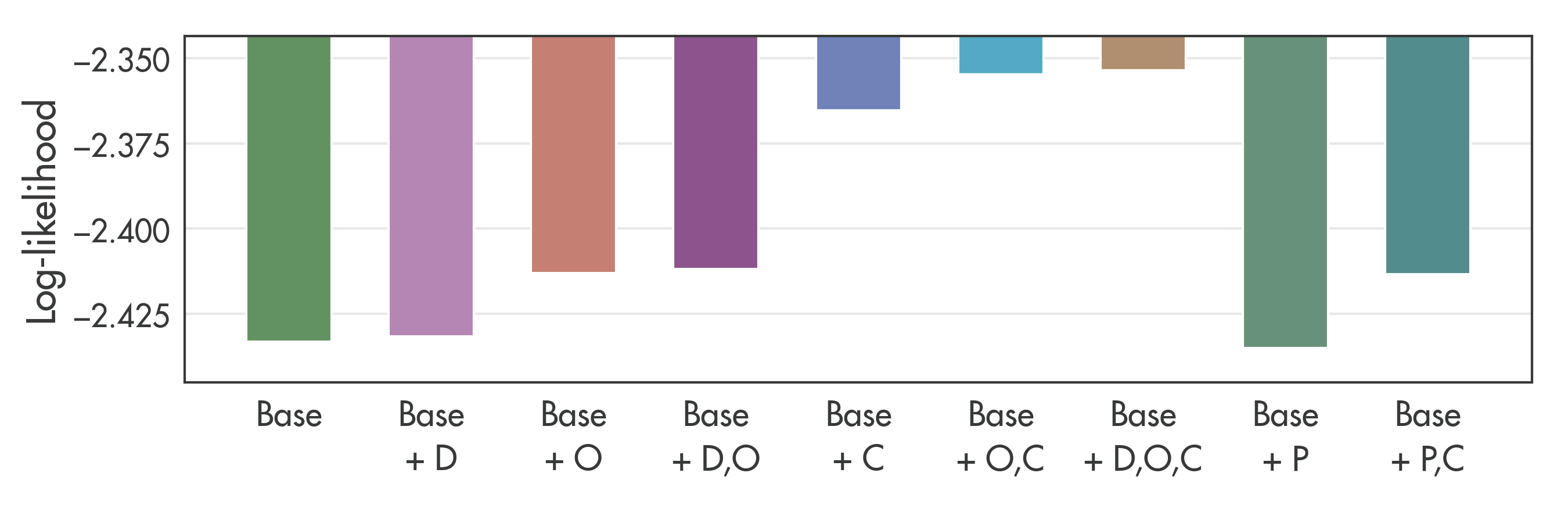}
\caption{\textit{Variants of Digital Representatives}. Mean log-likelihood of ground-truth critiques from human participants (from the validation set), evaluated under various digital representatives. All these DRs have 1B parameters and were fine-tuned on datasets conditioned on diverse additional information, as indicated on the x-axis (refer to Table \ref{tab:prompt-info} for details and examples). The optimal variant (\textit{Base+O+C}) incorporates participant $i$'s opinions and critiques to other questions. This variant performs very similarly to its counterpart that additionally includes demographic information (\textit{Base+D+O+C}). However, we opted for the former for simplicity. Note that variants based solely on demographics (\textit{Base+D} or position scoring (\textit{Base+P}) perform much worse. This suggests that integrating participant-specific few-shot information enhances both task- and self-consistency.}
\label{fig:dr-variants}
\end{figure}

\begin{figure}[H]
\centerline{
\includegraphics[width=0.5\columnwidth]{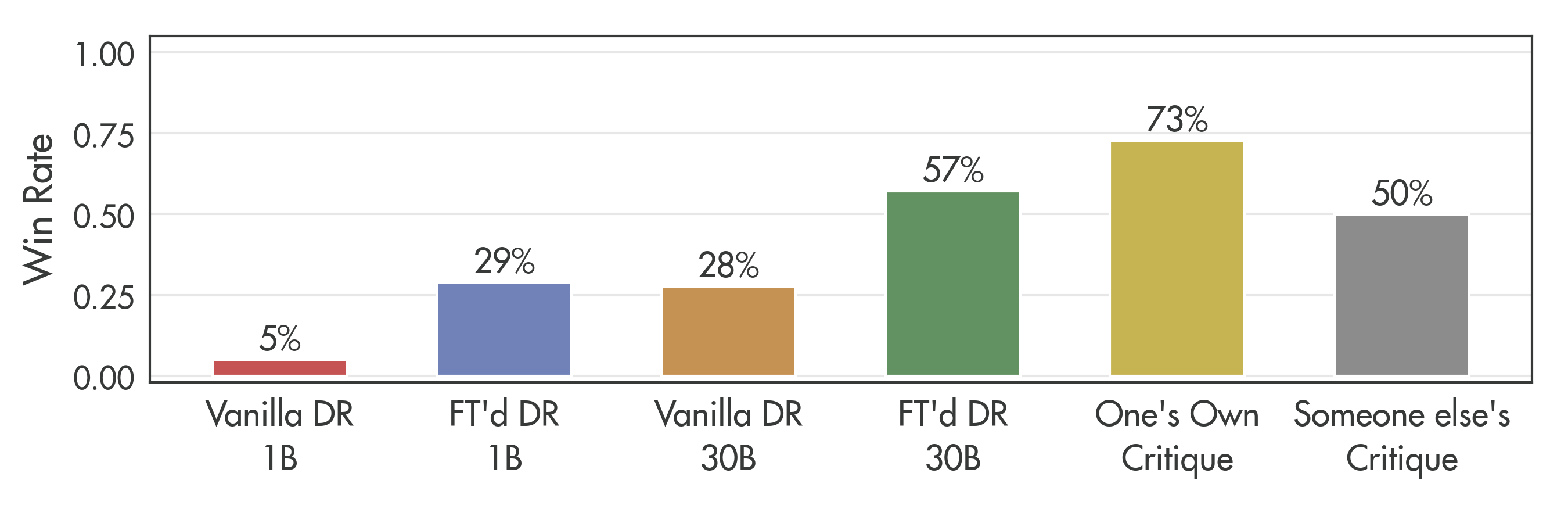}
\includegraphics[width=0.5\columnwidth]{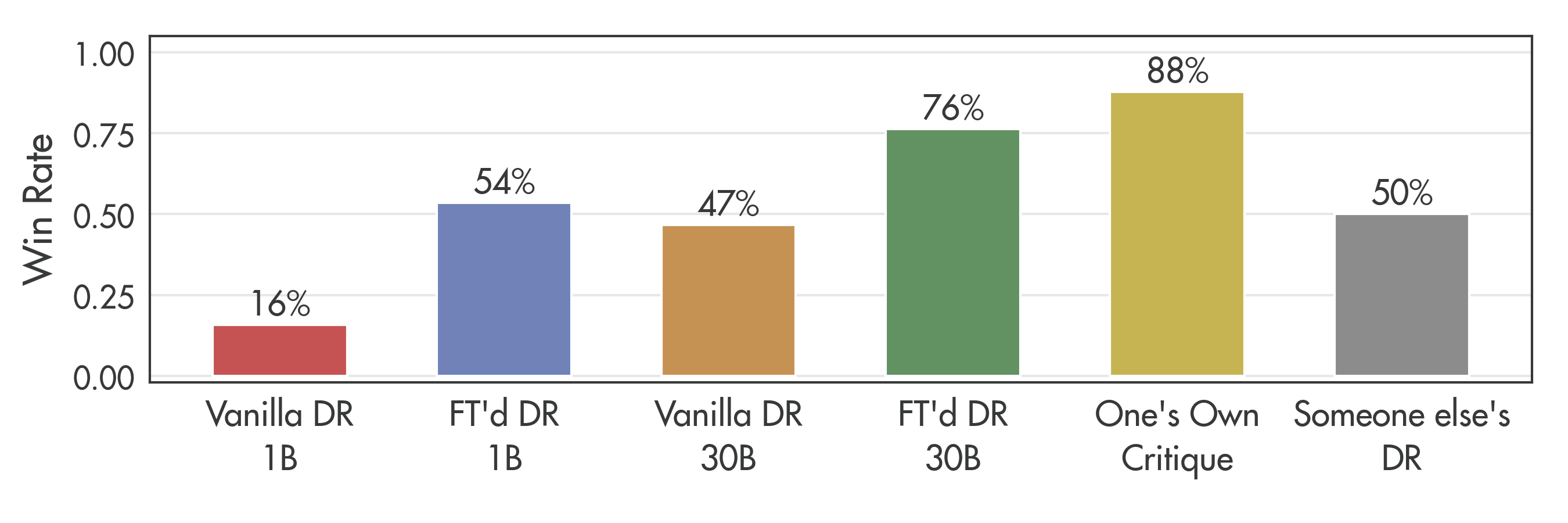}
}
\centerline{
\includegraphics[width=0.5\columnwidth]{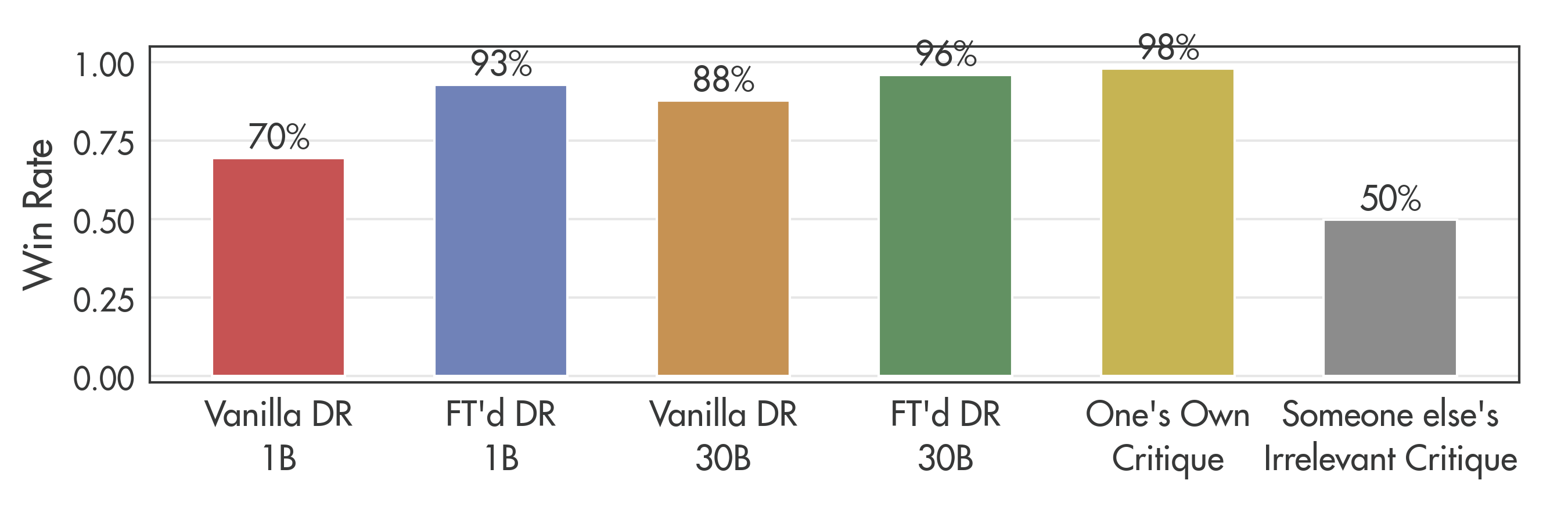}
\includegraphics[width=0.5\columnwidth]{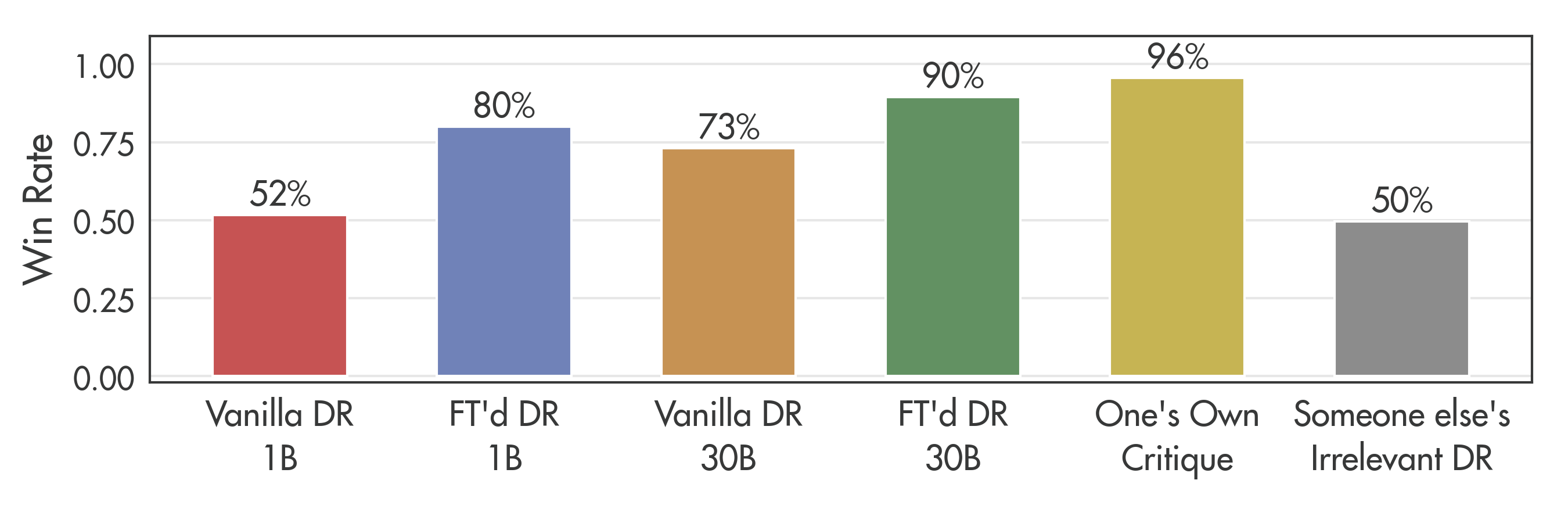}
}
\centerline{
\includegraphics[width=0.5\columnwidth]{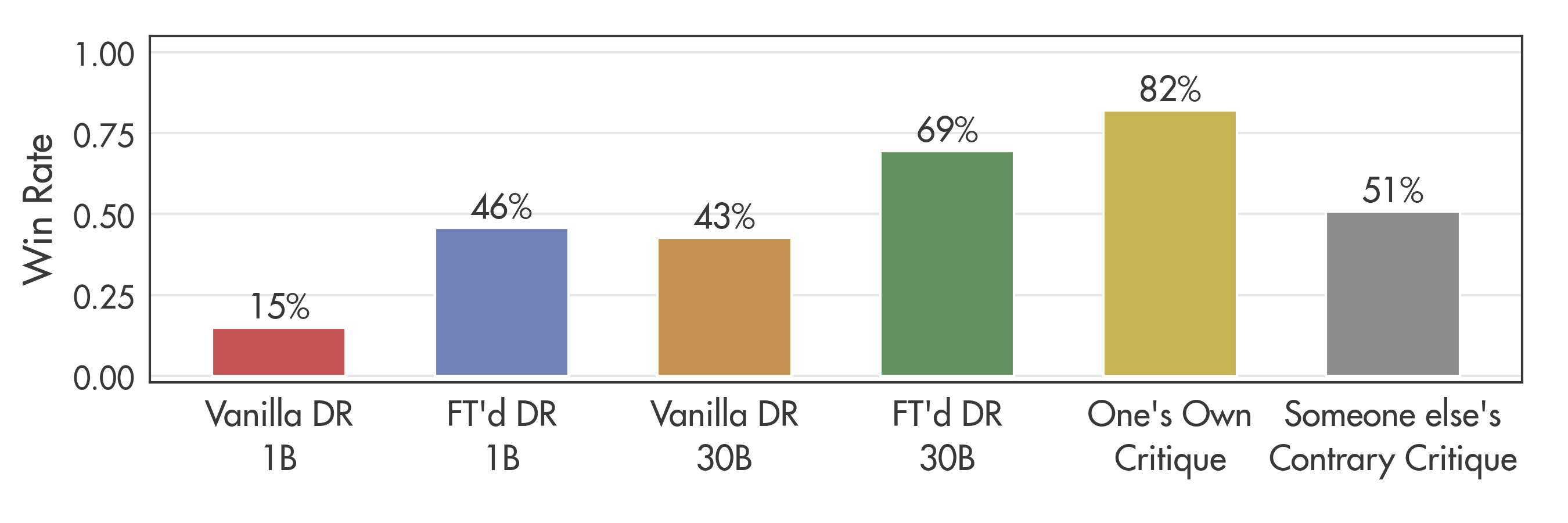}
\includegraphics[width=0.5\columnwidth]{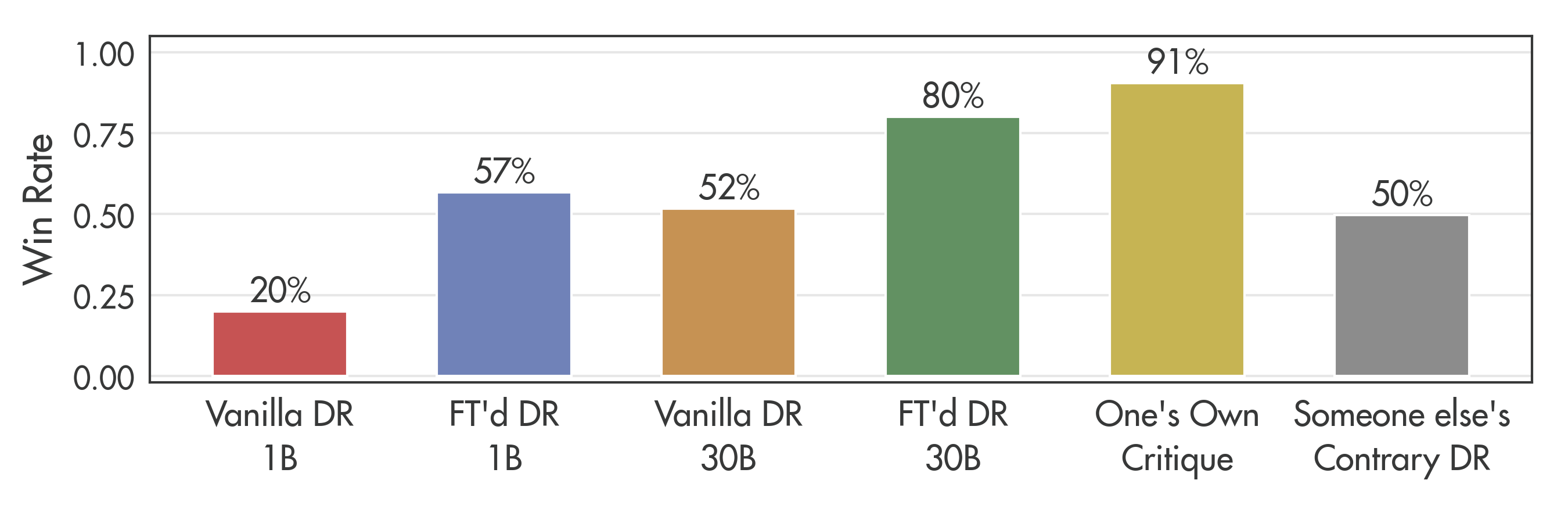}
}
\centerline{
\includegraphics[width=0.5\columnwidth]{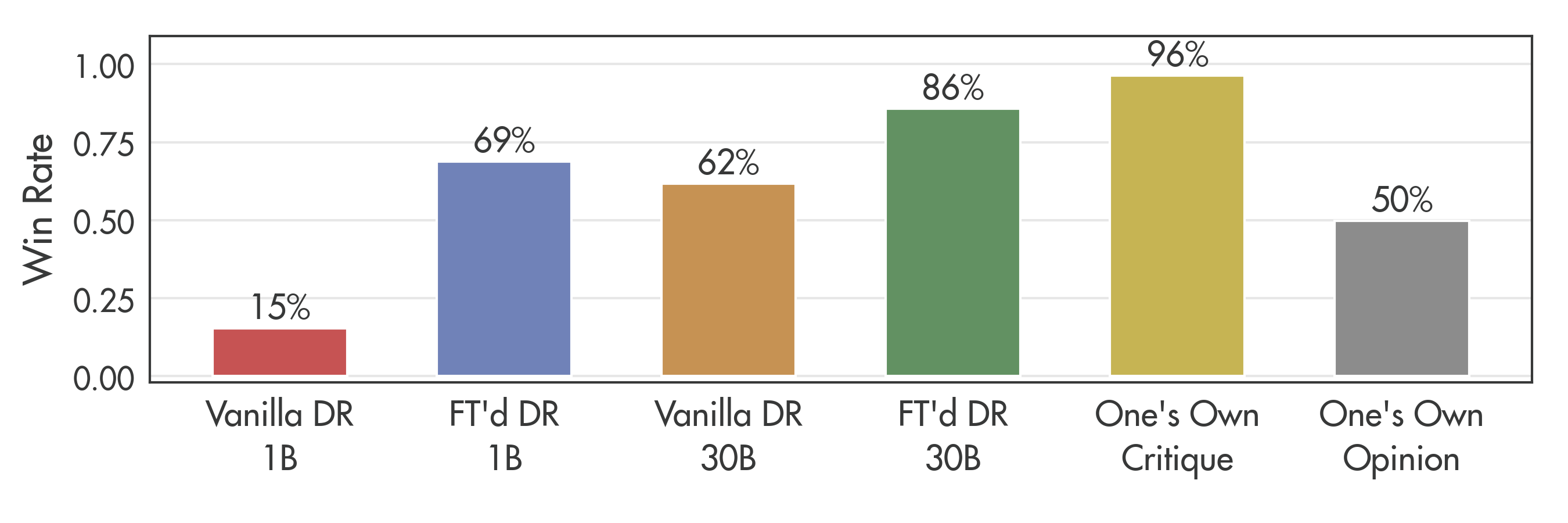}
\includegraphics[width=0.5\columnwidth]{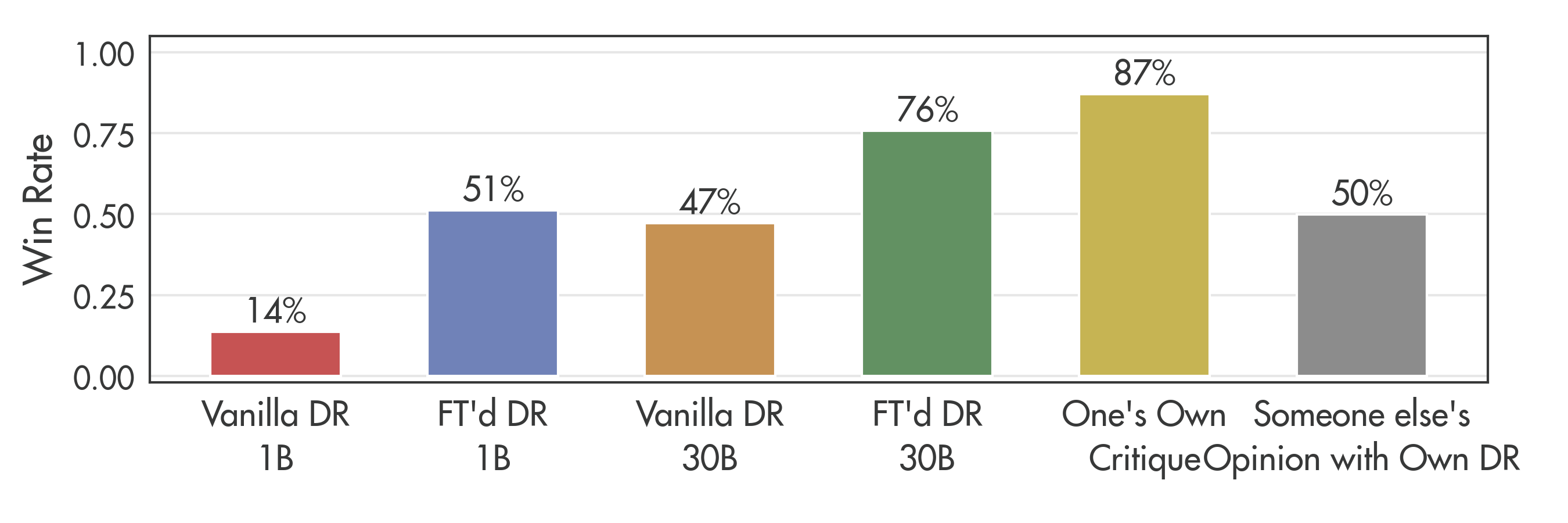}
}
\caption{\textit{Autorater Critique Evaluations: Ablations}. The autorater's win-rate of sampled critiques against a custom baseline, as specified by the grey bar within each plot (included as a sanity check, expected around 50\%). The ground truth (\textit{One's Own Critique}) is also included to set the performance ceiling. The main take-away is that the 30B fine-tuned DR (green) closely matches the performance of ground truth critiques (golden), suggesting its ability to learn both task-specific aspects (e.g., critique-ness, topic) and participant-specific nuances (e.g., individual preferences, style). A similar trend is observed for the 1B fine-tuned DR, albeit to a lesser extent.}
\label{fig:autoeval-critiques}
\end{figure}

~

\begin{story}[ht]
 \framedbox[0.95\textwidth]{
A group of participants is engaged in debates regarding important questions. They have been tasked with coming up with a consensus for a new debate question.

This participant has shared their opinions on other open debate questions, as outlined below.

\textbf{Question 1 from a past debate:} Should students be required to pass a literacy and numeracy test before graduating from primary school?

\textbf{Opinion of the participant about question 1:} I'm against this, children should not be under pressure at such a young age and should concentrate more on being children. Obviously monitor literacy and numeracy but to make them pass a test is unnecessary. Some people develop academically as they get older, so a test at this young age is meaningless. Children will have enough pressure as adults, let them enjoy being children.

\textbf{Critique of the participant about question 1:} Totally agree one hundred per cent with this, pupils should not have to pass a test before graduating but they should be monitored and helped if necessary. Let children be children.
Question 2 from a past debate: Should we require people to pay a fixed tax on any amount of wealth they own above £10 million?

\textbf{Opinion of the participant about question 2:} There is a growing wealth gap in the UK and the world and a tax like this would help narrow that gap, help the country pay its way and help to invest in important infrastructure. However we do not want to set this fixed amount of tax too high that it puts wealthy individuals off from living here, investing here and spending money. I do think though £10 million is the right figure to apply this tax.

\textbf{Critique of the participant about question 2:} Like the group opinion, it aligns well with what my response was. Totally agree with the money helping to fund the public sector and poorest in society. My concern was that the tax should not be set to high and 1\% seems a fair figure. That 1\% would still generate a substanial sum for the country.

Please offer a critique that accurately represents the participant's opinion on a new debate question in relation to the group draft consensus statement.

\textbf{Current question under debate:} Should the government spend more on science and technology research?

\textbf{Opinion of the participant about the question under debate:} I think its really important that the government spends more on Science and Technology. If you think about the problems we have as a society and as a planet - the answers lie with Science and Technology. We have just had a serious pandemic and we were helped by scientists coming up quickly with a vaccination. One of the ways out of the Climate Change crisis is through science and technology coming up with solutions as we as individuals do not want to change pur habits. More money going into Science and Technology will help us counter these crises and future problems. If we don't put the money in, the next pandemic maybe more severe and costly.

\textbf{Group draft consensus statement (to be critiqued):} The government should spend more on science and technology research as this would lead to better technology and medical discoveries to help the world.

\textbf{Critique of the participant:}}
 \caption[]{Prompt example for the digital representative used in the main text (\textit{Base+O+C}).}
 \label{box:prompt-example}
\end{story}

\begin{figure}[H]
\centering
\includegraphics[width=0.96\columnwidth]{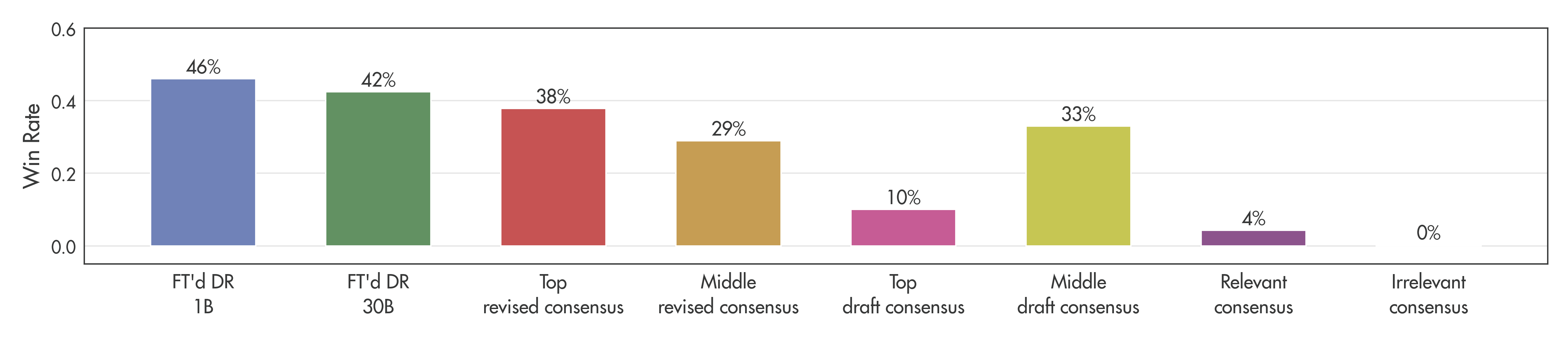}
\caption{\textit{Autorater Consensus Evaluations: Ablations}. The autorater's win-rate of generated consensuses (based on single participant substitutions with DRs) against the baseline (based solely on ground truth critiques). For comparison, various baseline consensus statements (as specified on the x-axis) are included from the original dataset. This ablation demonstrates that DRs contribute to generating consensuses that are competitively similar to human-selected consensuses (i.e. red bar).}
\label{fig:consensus-baselines}
\end{figure}

\end{document}